\documentclass[10pt,twocolumn,letterpaper]{article}
\usepackage{preprint}
\usepackage{times}
\usepackage[hyphens]{url}
\usepackage{graphicx}
\usepackage[numbers,sort&compress]{natbib}
\usepackage{booktabs}
\usepackage{array}
\usepackage{placeins}
\usepackage{amsmath,amssymb,amsfonts}
\usepackage{tikz}
\usetikzlibrary{arrows.meta,fit,positioning}
\usepackage[breaklinks=true,hidelinks,bookmarks=true]{hyperref}
\newcommand{\keywords}[1]{\par\noindent\parbox{\columnwidth}{\normalfont\textbf{\emph{Keywords: #1}}}\par}
\newcommand{\glm}{GLM-Image}
\newcommand{\qwen}{Qwen-Image}

\title{Four-Step Text-Bearing Image Generation via Distillation with Local Response Regularization and Teacher-Guided Training}
\newcommand{\PaperAuthorMetadata}{Kairui Wen, Sheng Ye, Yong-Jin Liu}
\author{%
  Kairui Wen\quad Sheng Ye\quad Yong-Jin Liu\textsuperscript{*}\\[5pt]
  {\small Tsinghua University, China}\\[4pt]
  {\small\href{mailto:wkr21@mails.tsinghua.edu.cn}{\nolinkurl{wkr21@mails.tsinghua.edu.cn}}\quad
  \href{mailto:yec22@mails.tsinghua.edu.cn}{\nolinkurl{yec22@mails.tsinghua.edu.cn}}\quad
  \href{mailto:liuyongjin@tsinghua.edu.cn}{\nolinkurl{liuyongjin@tsinghua.edu.cn}}}\\[4pt]
  {\small\textsuperscript{*}Corresponding author.}%
}

\makeatletter
\newcommand{\teasercaption}[1]{\begingroup\def\@captype{figure}\caption{#1}\endgroup}
\makeatother
\hypersetup{pdftitle={Four-Step Text-Bearing Image Generation via Distillation with Local Response Regularization and Teacher-Guided Training},pdfauthor={\PaperAuthorMetadata},pdfsubject={Research preprint},pdfkeywords={diffusion distillation, visual text rendering, text-to-image generation}}
\begin{document}
\twocolumn[{%
\begin{@twocolumnfalse}
\maketitle
\begingroup
\centering
\definecolor{tealresult}{HTML}{087F73}
\definecolor{amberresult}{HTML}{BD721C}
\setlength{\tabcolsep}{3pt}
\setlength{\fboxsep}{0pt}
\setlength{\fboxrule}{.65pt}
\begin{tabular}{@{}cccc@{}}
\includegraphics[width=.239\textwidth]{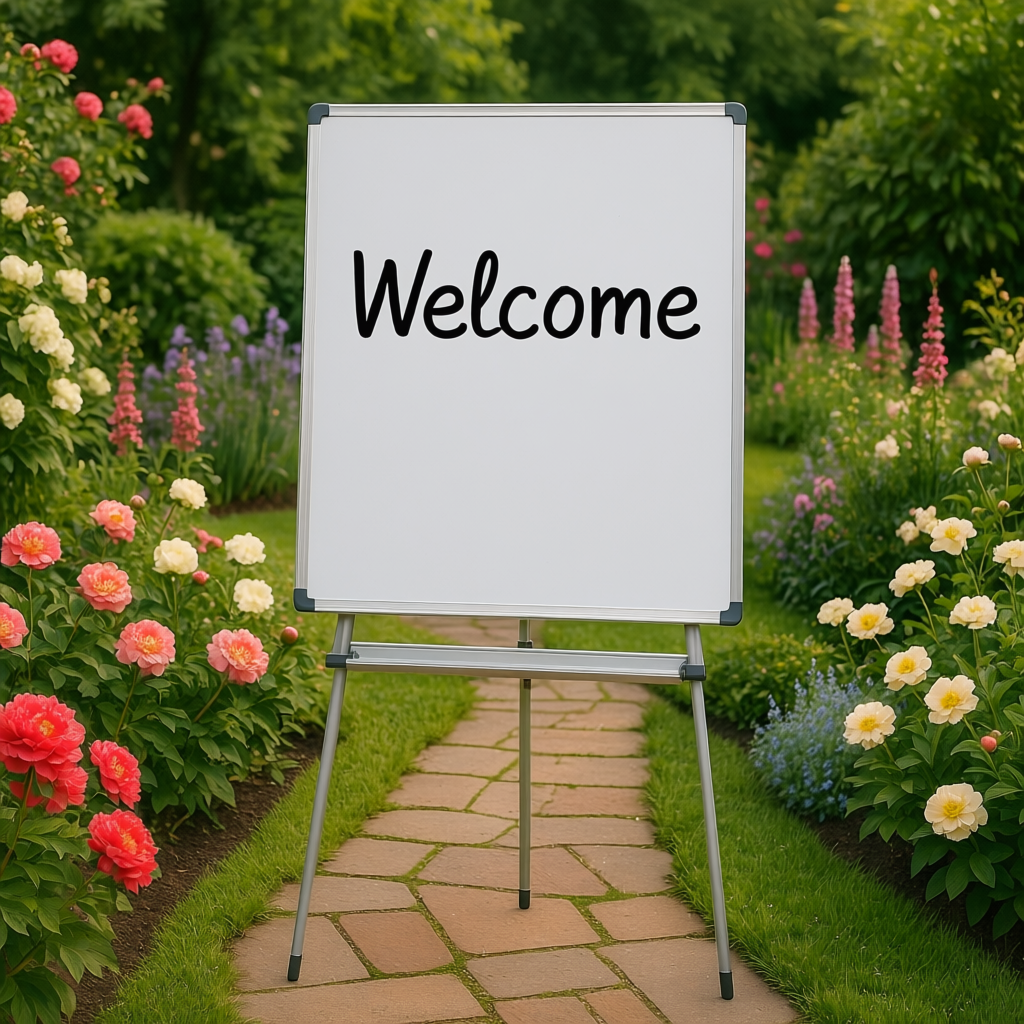} &
\includegraphics[width=.239\textwidth]{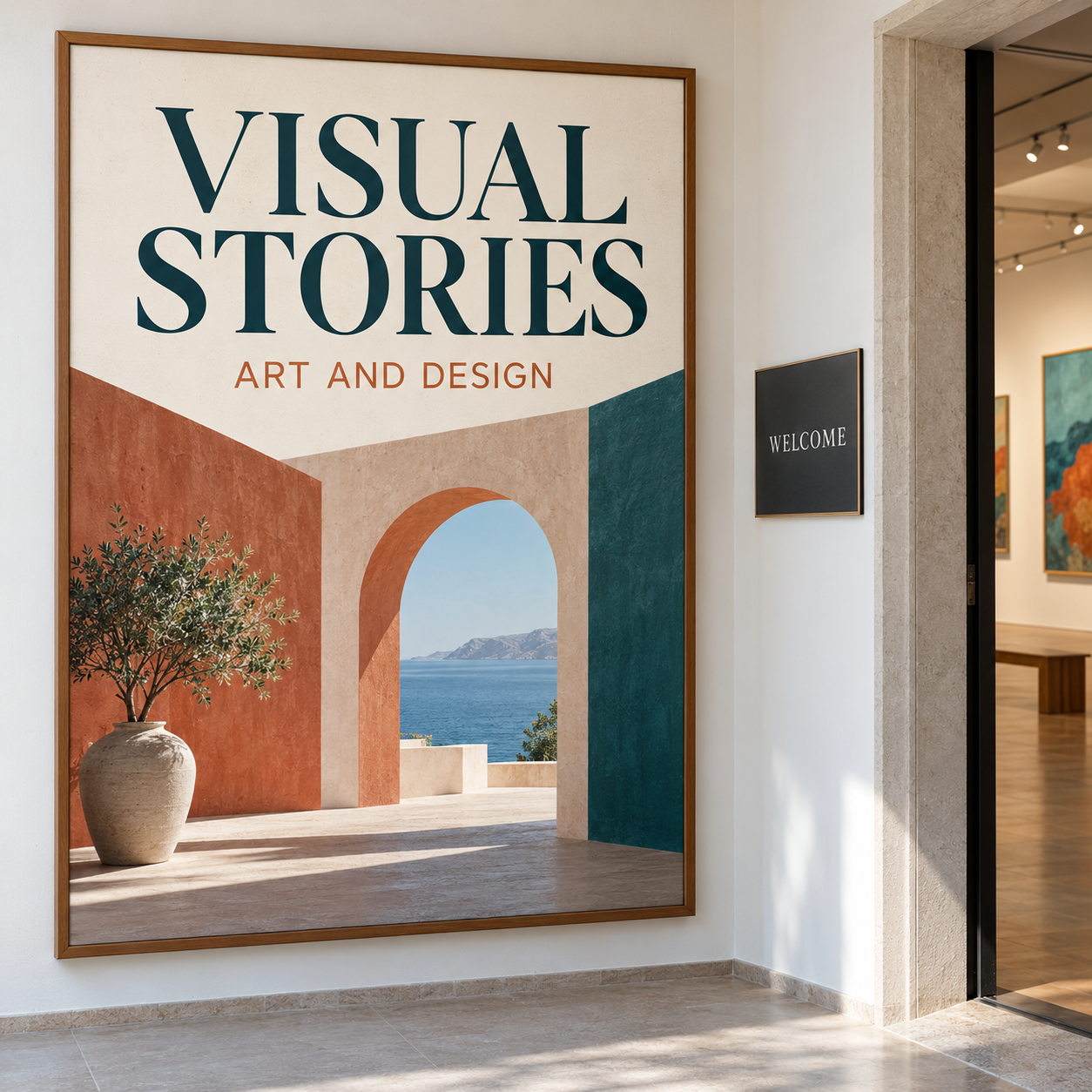} &
\includegraphics[width=.239\textwidth]{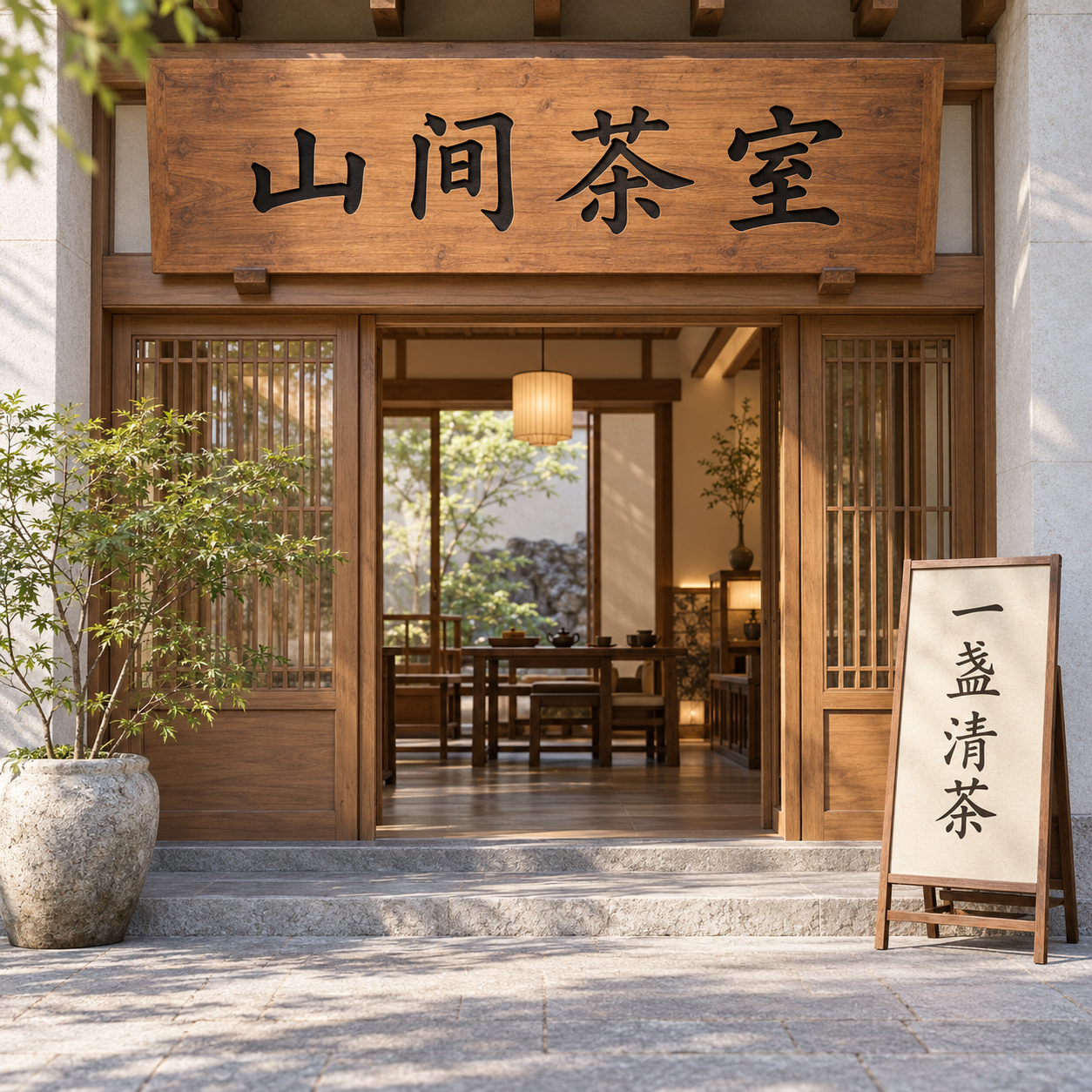} &
\includegraphics[width=.239\textwidth]{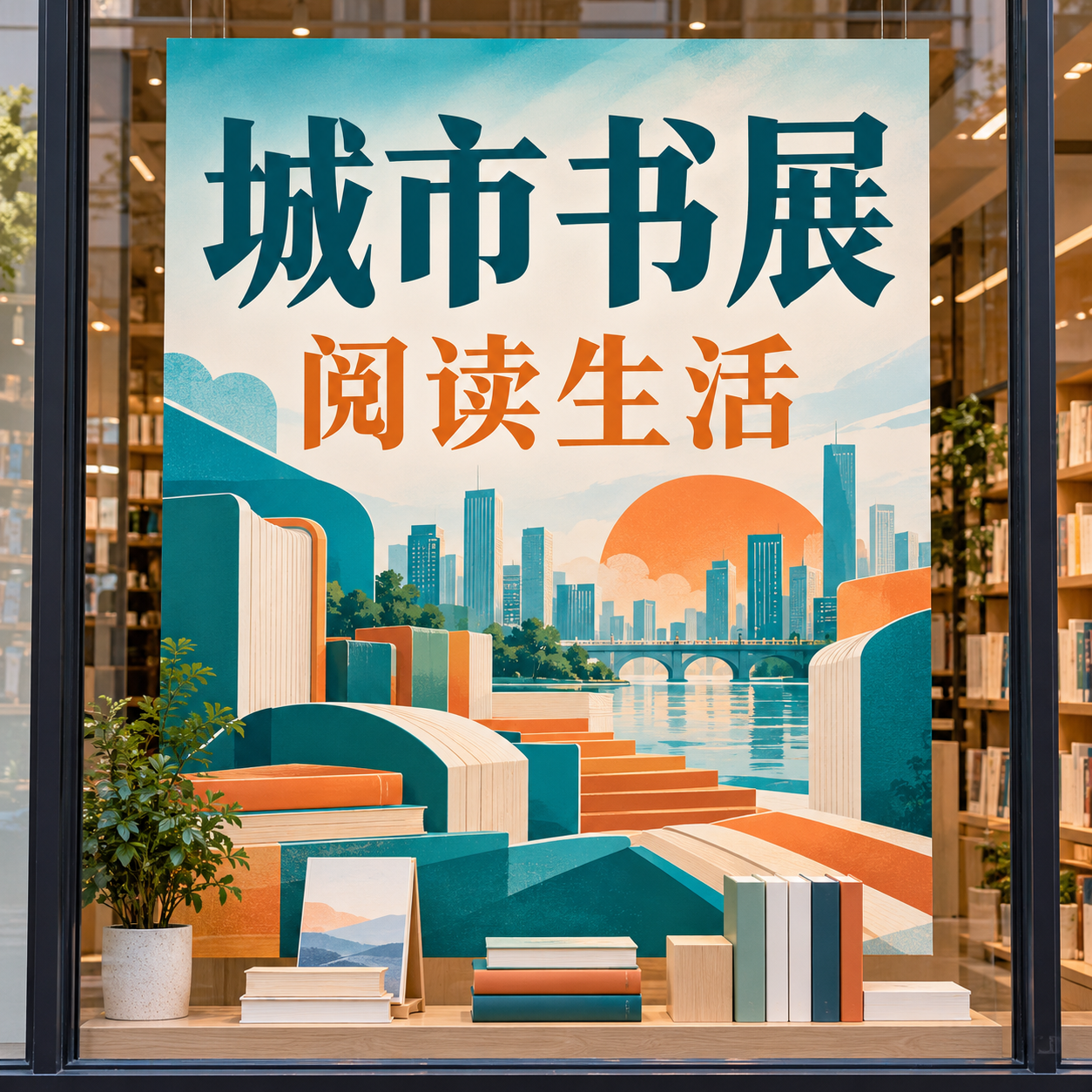}\\[2pt]
\small Garden signage & \small Exhibition poster & \small Tea-shop signage & \small Book-fair poster\\[2pt]
\multicolumn{4}{c}{\small\textbf{(a) Text-bearing visual media}}\\[5pt]
\fcolorbox{tealresult}{white}{\includegraphics[width=.236\textwidth,trim=280bp 619bp 244bp 215bp,clip]{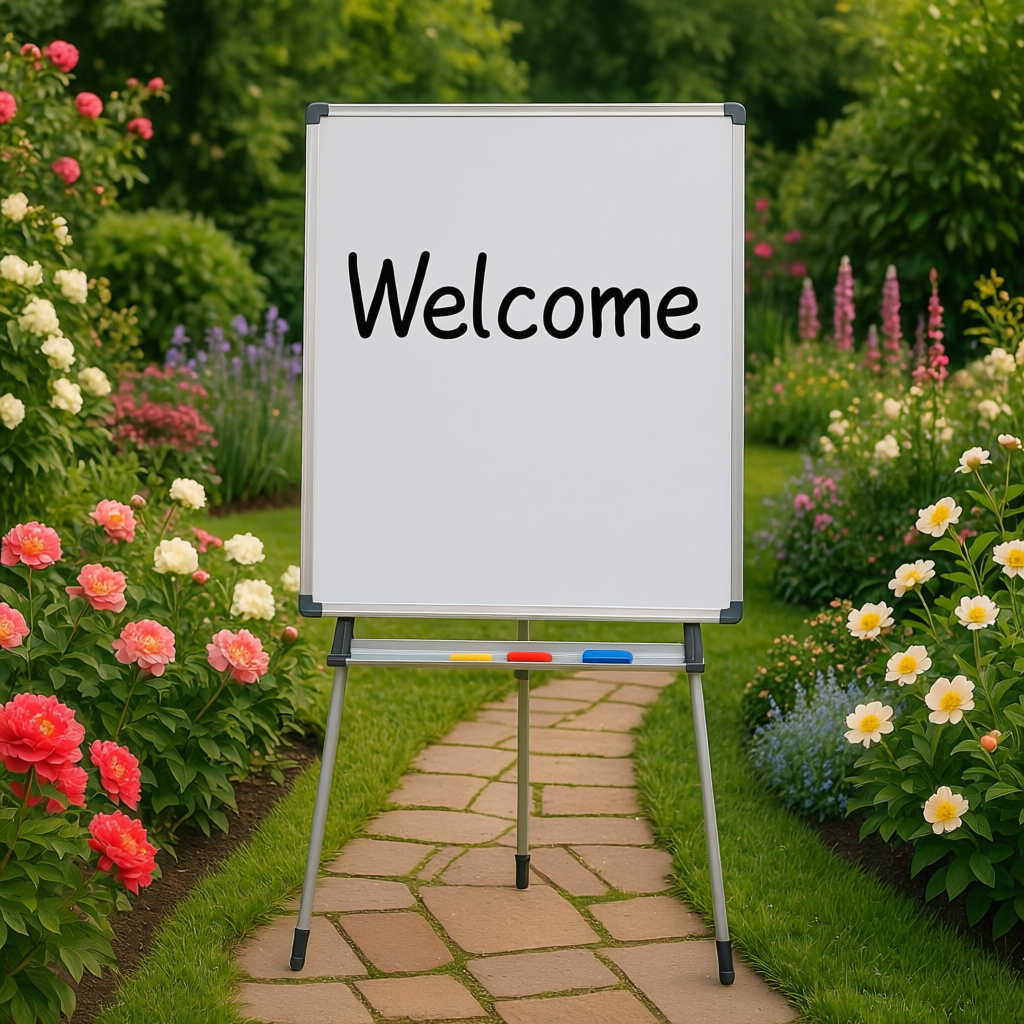}} &
\fcolorbox{amberresult}{white}{\includegraphics[width=.236\textwidth,trim=280bp 619bp 244bp 215bp,clip]{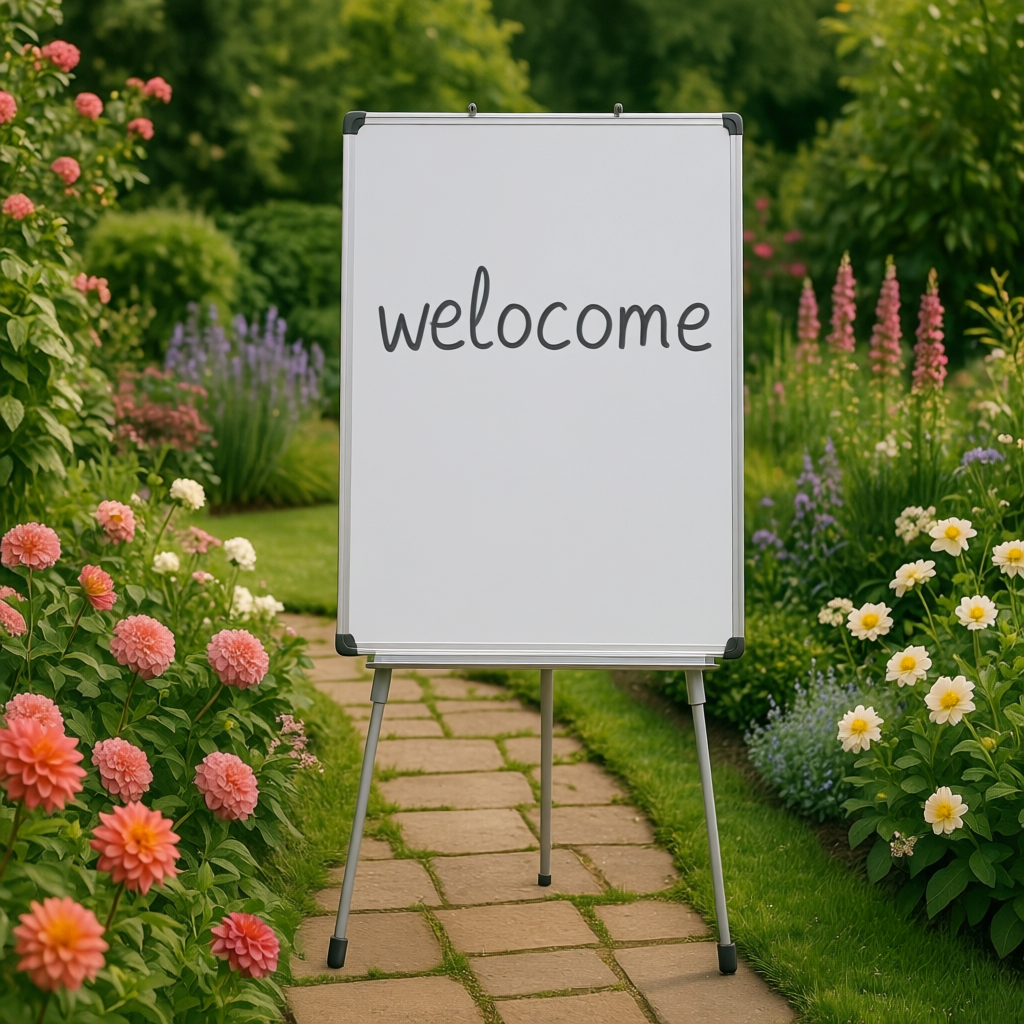}} &
\fcolorbox{amberresult}{white}{\includegraphics[width=.236\textwidth,trim=280bp 619bp 244bp 215bp,clip]{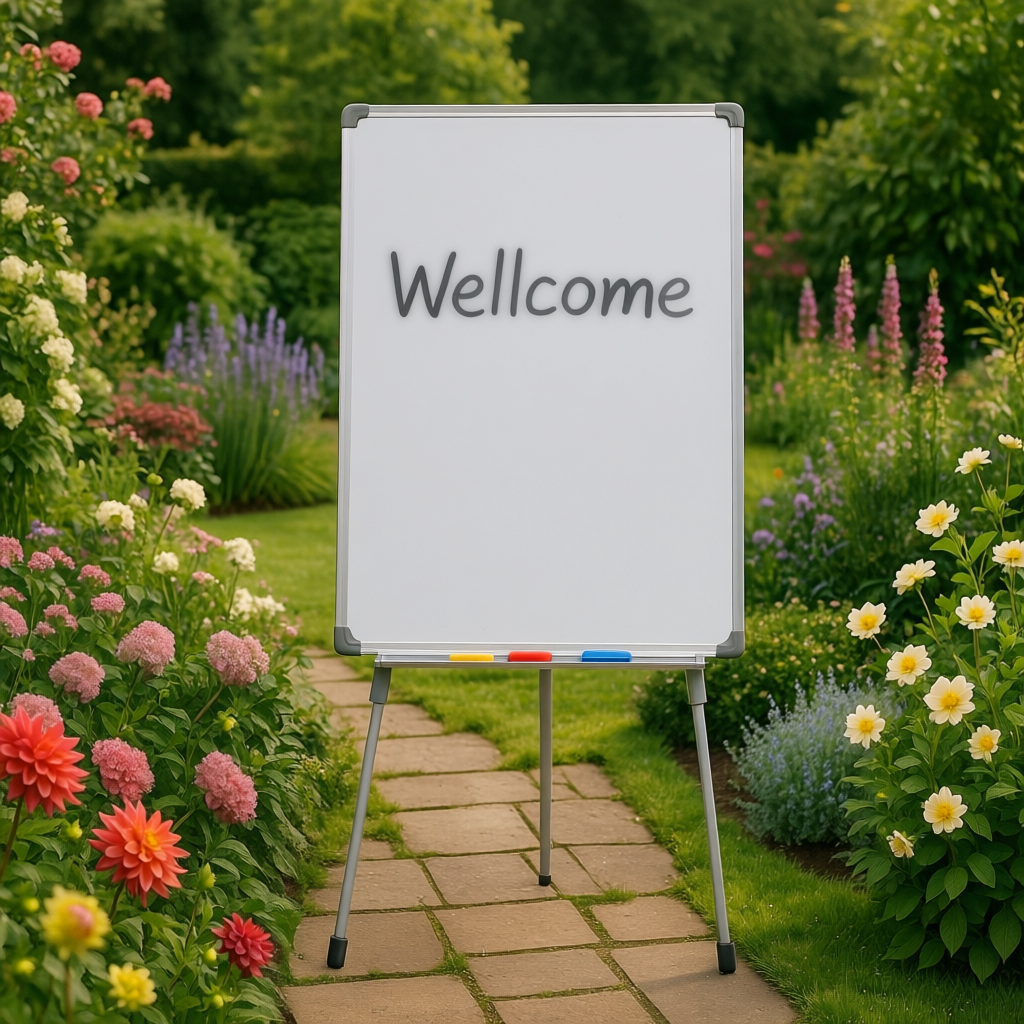}} &
\fcolorbox{tealresult}{white}{\includegraphics[width=.236\textwidth,trim=280bp 619bp 244bp 215bp,clip]{fig2_teacher_images.png}}\\[2pt]
\small Teacher: 50 steps & \small No text training: 4 steps & \small Synthetic text: 4 steps & \small Ours: 4 steps\\[2pt]
\multicolumn{4}{c}{\small\textbf{(b) Same garden prompt and seed: target string ``Welcome''}}\\[4pt]
\multicolumn{4}{c}{\footnotesize\textcolor{tealresult}{\textbf{4 vs.\ 50 steps: $12.5\times$ fewer denoiser evaluations}}
\quad\textbar\quad Four-step GenEval: \textbf{0.84} (ours) vs.\ 0.81 (TDM)}
\end{tabular}
\teasercaption{Four-step synthesis with preserved literal content. (a) Our ``Welcome'' garden scene alongside three application illustrations. (b) Matched experimental crops; teal and amber indicate correct and incorrect spelling. Our student recovers the target word with $12.5\times$ fewer denoiser evaluations than the teacher and improves GenEval over four-step TDM.\label{fig:teaser}}
\par
\endgroup

\vspace{6pt}
\end{@twocolumnfalse}
}]
\begin{abstract}
Generating text-bearing images requires coherent scenes and accurate characters. Large text-to-image models improve both capabilities through repeated denoising, at substantial sampling cost. Few-step distillation reduces this cost but requires the student to organize scenes and refine glyphs within a short trajectory.
We present a four-step distillation framework combining local response regularization with teacher-guided text training. Built on distribution matching, the framework distills the GLM-Image DiT while retaining its autoregressive visual prior, glyph encoder, and VAE. A path-length penalty promotes consistent local responses of the conditional denoising field. Progressive training combines synthetic strings with teacher-generated imagery to preserve character accuracy and scene-integrated typography. The student achieves GenEval Overall 0.84 and CVTG-2K word accuracy 0.5045, compared with 0.81 and 0.1690 for four-step TDM. A coefficient study favors $10^{-5}$, and Qwen-Image experiments show additional gains from response regularization and text supervision. Single-image inference takes 1.742\,s, compared with 5.818\,s for the teacher, yielding a $3.34\times$ speedup at the same 40\,GB GPU memory requirement.
\end{abstract}


\section{Introduction}

Text-to-image diffusion models support diverse visual synthesis for design and content creation~\citep{rombach2022ldm,wu2025qwenimage}. Text-bearing images, such as posters, signs, and information panels, place an additional demand on these models: the generated scene must communicate specific written content through legible, correctly positioned characters~\citep{du2025textcrafter,wu2025qwenimage}. Their synthesis therefore requires both scene-level composition and precise local structure. A small character error can alter the intended message even when the surrounding image appears plausible.

Few-step distillation reduces sampling cost by replacing a long denoising trajectory with a small number of learned transitions. Compressing a fifty-step teacher into a four-step student while retaining its visual prior and conditioning interface supports rapid exploration of text-bearing designs.
Diffusion models build image quality through iterative refinement~\citep{sohl2015nonequilibrium,ho2020ddpm,song2021sde,rombach2022ldm}. Distillation accelerates this process through progressive compression, consistency learning, adversarial supervision, or distribution matching~\citep{salimans2022progressive,song2023consistency,sauer2024add,yin2024dmd}. DMD2 extends distribution-matching distillation (DMD) to multi-step students through improved fake-score tracking and inference-state simulation~\citep{yin2024dmd2}. Trajectory Distribution Matching (TDM) and Phased DMD further improve trajectory supervision and multi-step optimization~\citep{luo2025tdm,fan2026phaseddmd}. The resulting short trajectories must retain the compositional and textual capabilities of the pretrained teacher.

Four-step distillation imposes two related learning demands. Each update must organize object relations while resolving the fine strokes and spacing of requested text. Distribution matching supplies a broad visual target; explicit strings and layouts supply the local constraints needed for literal rendering. We study this setting with \glm{}~\citep{zai2026glmimage}, whose autoregressive visual prior and glyph conditioning provide a natural interface between scene planning and continuous DiT refinement. We retain these conditioning paths and concentrate distillation on the denoiser.

Our framework combines a response prior with teacher-guided text supervision. We adapt path-length regularization to the conditional denoising field, measuring latent-to-output responses while holding the visual prior and glyph features fixed. The adaptive target promotes consistent response magnitudes across training states, providing a regularity prior for coarse denoising transitions. Teacher-guided training complements this prior with synthetic examples that specify exact strings and teacher-generated images that integrate typography into scenes. Text complexity progresses from short strings to dense layouts.

We evaluate composition on GenEval~\citep{ghosh2023geneval} and literal rendering on CVTG-2K~\citep{du2025textcrafter}. Response regularization raises the four-step GLM student from 0.81 to 0.84 GenEval Overall. Teacher-image mixing increases word accuracy from 0.2086 to 0.5045 over synthetic-only training. The complete framework also outperforms four-step TDM on both benchmarks. A coefficient sweep favors $10^{-5}$, and a second study with \qwen{} evaluates response regularization alongside a teacher-specific text-training procedure. In single-image inference, the GLM student reduces latency from 5.818\,s to 1.742\,s while both models require 40\,GB of GPU memory.

In summary, this paper makes three contributions:\nopagebreak[4]
\begin{itemize}
\item We present a four-step framework for text-bearing image synthesis that retains the pretrained semantic and glyph-conditioning interface of \glm{}.
\item We adapt path-length regularization to the conditional denoiser and evaluate its effect on composition and text detail across two DiT teacher families.
\item We develop teacher-guided text training that combines explicit strings with scene-integrated typography and improves literal rendering across multiple text-region counts.
\end{itemize}

\section{Related Work}
Our framework connects efficient image synthesis, diffusion distillation, local response regularization, and visual text generation. We review these areas and distinguish our response penalty from related forms of differential supervision.

\paragraph{Efficient image synthesis.}
Latent diffusion enables high-resolution synthesis in a compressed representation~\citep{rombach2022ldm}, and DiT scales denoising with transformers~\citep{peebles2023dit}. CLEAR, EDiT, LiT, and contiguous-layer pruning reduce computation within individual model evaluations~\citep{liu2025clear,becker2025edit,wang2025lit,ma2026ppcl}. Bidirectional Multiway Transformers combine visual reconstruction with modality-specific computation~\citep{bao2025multiway}. Our framework reduces the number of denoising evaluations while retaining the teacher architecture and conditioning paths.

\paragraph{Few-step generation and distillation.}
Progressive Distillation, Consistency Models, and Adversarial Diffusion Distillation accelerate synthesis through trajectory compression, self-consistency, and adversarial supervision~\citep{salimans2022progressive,song2023consistency,sauer2024add}. Shortcut Models and MeanFlow learn short-step or interval-average dynamics~\citep{frans2025shortcut,geng2025meanflow}. Continuous-time consistency models improve differential training, and rCM combines this formulation with score distillation at large scale~\citep{lu2025scm,zheng2026rcm}.

DMD2 improves online score tracking and introduces inference-state simulation for multi-step students~\citep{yin2024dmd,yin2024dmd2}. TDM aligns trajectory distributions with step-aware supervision, and autoregressive distillation uses trajectory history~\citep{luo2025tdm,kim2025ard}. Phased DMD organizes score matching and model capacity into subintervals~\citep{fan2026phaseddmd}. SiD extends score-identity distillation to flow-matching DiTs~\citep{zhou2025sidflow}; SMD uses on-policy velocity regression with a training-only residual proxy~\citep{zhang2026smd}. Distillation also supports fractional evaluation budgets through caching~\citep{li2026fractional}. Our added response prior acts on local denoiser behavior and complements the distributional training objective with dedicated text supervision.

\paragraph{Differential supervision and local response regularization.}
StyleGAN2 promotes well-conditioned style-to-image mappings through path-length regularization~\citep{karras2020stylegan2}. We apply its adaptive response-magnitude principle to a time-conditioned denoising field with fixed semantic and glyph features. GAD aligns teacher and student directional derivatives at initial or intermediate noisy states~\citep{huang2026gad}. Our target is a running scalar response scale, which regularizes the magnitude profile of the student field. The supervised mapping and target distinguish these uses of Jacobian products. Table~\ref{tab:method_positioning} compares the training mechanisms, and Section~\ref{sec:response} defines our objective.

\begin{table*}[t]
\centering\footnotesize
\setlength{\tabcolsep}{4pt}
\renewcommand{\arraystretch}{1.08}
\caption{Training objectives of the methods most relevant to our work. Our framework integrates a conditional response-magnitude prior with teacher-guided text supervision while retaining the distribution-matching backbone and the teacher's conditioning interface.}
\label{tab:method_positioning}
\begin{tabular}{@{}p{.15\textwidth}p{.36\textwidth}p{.44\textwidth}@{}}
\toprule
Method & Main supervision & Relation to the proposed response prior\\
\midrule
DMD2~\citep{yin2024dmd2} & Teacher/fake-score distribution matching; inference-state simulation & Supplies the base distributional supervision complemented by our conditional response term.\\
TDM~\citep{luo2025tdm} & Step-aware trajectory distribution matching & Supervises trajectory distributions; our added term shapes local input responses.\\
Phased DMD~\citep{fan2026phaseddmd} & Progressive subinterval distribution and score matching & Allocates supervision and capacity by phase; our added term acts on a shared conditional denoiser.\\
rCM~\citep{zheng2026rcm} & Continuous-time consistency plus score distillation & Uses differential consistency and teacher scores; our VJP measures response magnitude at a fixed condition and time.\\
GAD~\citep{huang2026gad} & Teacher--student directional derivative alignment & Matches response vectors to a teacher target; our target is an adaptive scalar response scale.\\
StyleGAN2~\citep{karras2020stylegan2} & Style-to-image path-length regularization & Supplies the regularization principle; we apply it to the latent-to-denoising field before guidance.\\
Ours & Distribution matching, adaptive response magnitudes, and teacher-guided text data & Couples conditional response consistency with literal-text supervision for four-step synthesis.\\
\bottomrule
\end{tabular}
\end{table*}

\paragraph{Visual text generation.}
Text-bearing synthesis combines scene composition, character identity, and spatial layout. Seedream 2.0 uses bilingual modeling and glyph-aligned text representations, while Qwen-Image employs dedicated text data and staged training~\citep{gong2025seedream,wu2025qwenimage}. GLM-Image couples an autoregressive visual prior with a glyph-conditioned diffusion decoder~\citep{zai2026glmimage}. X2I transfers multimodal understanding into a DiT through attention distillation~\citep{ma2025x2i}. At the application level, PosterMaker combines character-discriminative text control with subject-preserving scene generation, and SimplePoster uses character-level position encoding for product-poster synthesis~\citep{gao2025postermaker,cui2026simpleposter}. AMO improves rendering through attention-modulated overshooting at sampling time~\citep{hu2025amo}; TextPecker addresses character-structure errors through perceptive reward learning~\citep{zhu2026textpecker}. We examine how to preserve these capabilities during distillation, using synthetic examples for explicit strings and teacher-generated images for integrated scene and typography supervision.

\paragraph{Controllable visual media.}
Controllable media synthesis combines precise conditioning with local structure preservation. Image-to-image diffusion supports semantic scene synthesis~\citep{liu2025iidm}; language-guided fashion editing and prompt-embedding inversion support controlled visual modification~\citep{wen2025lucie,li2026stylediffusion}. Procedural texturing, feature-enhanced diffusion, and diffusion-based sketch extraction extend this goal to materials and geometric structure~\citep{dang2025texpro,song2026fednet,yun2026diffsketch}. We address content preservation under a short denoising budget, using scene composition and literal text as complementary fidelity criteria.

\paragraph{Evaluating composition and literal fidelity.}
GenEval evaluates object-level compositional alignment~\citep{ghosh2023geneval}, while TextCrafter introduces CVTG-2K for multiple text regions in complex scenes~\citep{du2025textcrafter}. TextInVision examines text complexity and character preservation through VAE reconstruction~\citep{fallah2025textinvision}; STRICT tests text length, legibility, and instruction adherence~\citep{zhang2025strict}. We pair GenEval with CVTG-2K and inspect accuracy by region count to evaluate both scene organization and literal rendering.

\section{Problem Formulation and Background}
\label{sec:formulation}
\subsection{Conditional Synthesis of Text-Bearing Images}
Let $c$ specify scene content and literal strings. With a fixed decoder $D$ and conditioning state $\kappa(c)$ containing visual-prior tokens and glyph features, the student generator is
\begin{equation}
I_\theta(z,c)=D\!\left(\Psi_{\theta,1}\circ\cdots\circ
\Psi_{\theta,K}(z;\kappa(c))\right),
\label{eq:conditional_generator}
\end{equation}
where $z$ is Gaussian noise and $\Psi_{\theta,k}$ is one scheduler transition. We use $K=4$ and retain the teacher's conditioning modules and decoder. The objective is to preserve scene composition and the literal strings specified by the prompt within this reduced sampling budget.

\subsection{Noising and Score Parameterization}
For a clean latent $x_0$, the forward perturbation is
\begin{equation}
x_t=\alpha_t x_0+\sigma_t\epsilon,\qquad
\epsilon\sim\mathcal N(0,I).
\label{eq:forward_noise}
\end{equation}
The conditional score is $s(x_t,t,c)=\nabla_{x_t}\log p_t(x_t\mid c)$. Teacher and fake-score predictions are converted to this common parameterization before taking their difference. A noise predictor gives $s=-\widehat\epsilon/\sigma_t$. For linear flow interpolation $x_t=(1-t)x_0+t\epsilon$ with velocity $v=\epsilon-x_0$,
\begin{align}
\widehat x_0&=x_t-t\widehat v,\notag\\
\widehat\epsilon&=x_t+(1-t)\widehat v.
\label{eq:flow_conversion}
\end{align}
The scheduler determines conversion and timestep interpretation. Our response penalty is defined on the conditional denoising output, with a fixed dimensional normalization.

\subsection{Distribution Matching with an Online Score Model}
Distribution matching compares noised student and teacher distributions through
\begin{equation}
\mathcal D(\theta)=\mathbb E_t\!\left[
w(t)D_{\mathrm{KL}}(p_{\theta,t}\|p_{T,t})\right].
\label{eq:distribution_objective}
\end{equation}
The teacher supplies a reference score, while an online fake-score model tracks the evolving student distribution~\citep{yin2024dmd,yin2024dmd2}. DMD2 further supports multi-step students through inference-state simulation. We retain this framework and add conditional response regularization and text-data construction for capability preservation.

\section{Method}
We introduce the distribution-matching backbone and hybrid conditioning interface in Sections~\ref{subsec:distribution-matching}--\ref{subsec:hybrid-interface}. Sections~\ref{subsec:training-objective}--\ref{sec:normalized_response} define the joint objective and normalized response penalty. Sections~\ref{sec:text_training}--\ref{sec:text_construction} describe progressive text supervision and stream composition.

\subsection{Distribution-Matching Distillation Backbone}
\label{subsec:distribution-matching}
Let $T$ be a pretrained teacher and $S_\theta$ its four-step student conditioned on prompt $c$. Following DMD2~\citep{yin2024dmd2}, a frozen teacher supplies the reference score and an online fake-score model tracks the student distribution. For a noised sample $x_t$, the distribution-matching generator gradient is
\begin{equation}
\nabla_\theta\mathcal{L}_{\mathrm{DM}}\propto
-\mathbb{E}\left[w(t)\,\Delta s_t\,
\frac{\partial S_\theta(z,c)}{\partial\theta}\right],
\label{eq:dm_grad}
\end{equation}
where $\Delta s_t=s_T(x_t,t,c)-s_F(x_t,t,c)$ compares scores in a common noise parameterization, and $w(t)$ includes noise-level weighting. The fake-score branch receives four updates per generator update. The student uses the timestep labels 1000, 750, 500, and 250; their conversion to model-input indices and scheduler noise levels follows the teacher-specific interface.

\subsection{Hybrid AR--DiT Distillation Interface}
\label{subsec:hybrid-interface}
\glm{} combines an autoregressive visual-prior module with a DiT denoiser. We retain the prior and glyph-conditioning paths, and optimize the continuous refinement mapping. The autoregressive module $A_\phi$ generates prior tokens
\begin{equation}
p_{\mathrm{AR}}=A_\phi(c),
\label{eq:ar_prior}
\end{equation}
where $c$ is the text prompt. For text-to-image generation, the generated prior-token IDs are extracted from the AR output and upsampled to the target DiT grid before being supplied through the prior-token-id input. In parallel, text spans extracted from the prompt are encoded by a glyph encoder $E_{\mathrm{glyph}}$:
\begin{equation}
h_{\mathrm{glyph}}=E_{\mathrm{glyph}}(\operatorname{glyph}(c)).
\label{eq:glyph_condition}
\end{equation}
The student DiT therefore receives the continuous latent, the glyph embedding, the AR-derived prior, and the current timestep:
\begin{equation}
v_\theta=F_\theta(x_t,h_{\mathrm{glyph}},p_{\mathrm{AR}},t).
\label{eq:dit_interface}
\end{equation}

The AR prior supplies coarse semantic and spatial structure; glyph features specify the characters to render. Keeping both fixed preserves this division of labor while the student learns a shorter denoising trajectory.

The Flow Matching scheduler advances the continuous latent using the predicted velocity:
\begin{equation}
x_{\mathrm{next}}=\operatorname{Step}_t(v_\theta,x_t).
\label{eq:dit_step}
\end{equation}
This interface reuses the pretrained semantic and character representations at every denoising transition.

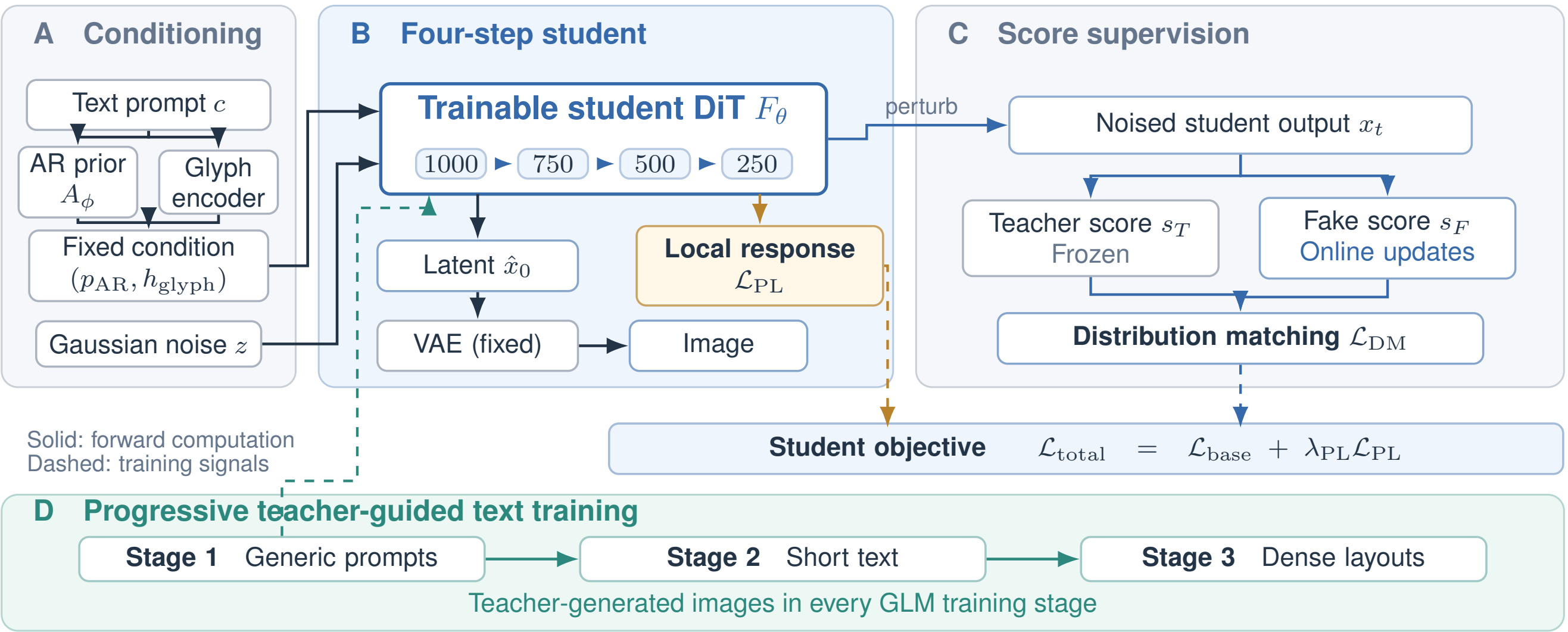
\begin{figure*}[t]
\centering
\resizebox{\textwidth}{!}{\begin{tikzpicture}[x=1cm,y=1cm,
  font=\sffamily\fontsize{8}{9.5}\selectfont,
  >=Latex,
  every node/.style={align=center,font=\sffamily\fontsize{8}{9.5}\selectfont},
  panel/.style={rounded corners=5pt,line width=.55pt},
  box/.style={rounded corners=3pt,line width=.65pt,inner sep=4pt},
  inference/.style={->,draw=ink,line width=.85pt},
  training/.style={->,draw=teal,dashed,line width=.8pt},
  scorepath/.style={->,draw=blue,line width=.8pt},
  heading/.style={anchor=west,font=\sffamily\bfseries\fontsize{9}{11}\selectfont},
  label/.style={font=\sffamily\fontsize{7}{8}\selectfont,text=muted}]
\definecolor{ink}{HTML}{243447}
\definecolor{muted}{HTML}{65758B}
\definecolor{blue}{HTML}{3569AA}
\definecolor{bluewash}{HTML}{EFF5FC}
\definecolor{teal}{HTML}{2C887D}
\definecolor{tealwash}{HTML}{EDF7F4}
\definecolor{amber}{HTML}{B8832C}
\definecolor{amberwash}{HTML}{FFF8E9}
\definecolor{graywash}{HTML}{F5F7FA}
\color{ink}

\draw[panel,draw=muted!35,fill=graywash] (0,2.65) rectangle (3.2,6.80);
\draw[panel,draw=blue!35,fill=bluewash] (3.45,2.65) rectangle (9.70,6.80);
\draw[panel,draw=muted!35,fill=graywash] (9.95,2.65) rectangle (17.0,6.80);
\node[heading,text=muted] at (.22,6.47) {A\quad Conditioning};
\node[heading,text=blue] at (3.67,6.47) {B\quad Four-step student};
\node[heading,text=muted] at (10.17,6.47) {C\quad Score supervision};

\node[box,draw=muted!55,fill=white,minimum width=2.65cm,minimum height=.46cm] (prompt) at (1.60,5.72) {Text prompt $c$};
\node[box,draw=muted!55,fill=white,text width=1.15cm,minimum height=.65cm,inner sep=2pt] (ar) at (.83,4.88) {AR prior\\$A_\phi$};
\node[box,draw=muted!55,fill=white,text width=1.15cm,minimum height=.65cm,inner sep=2pt] (glyph) at (2.36,4.88) {Glyph\\encoder};
\node[box,draw=muted!55,fill=white,minimum width=2.60cm,minimum height=.55cm,inner sep=2pt] (condition) at (1.60,3.97) {Fixed condition\\$(p_{\mathrm{AR}},h_{\mathrm{glyph}})$};
\node[box,draw=muted!55,fill=white,minimum width=2.10cm,minimum height=.42cm] (noise) at (1.60,3.12) {Gaussian noise $z$};
\draw[inference] (prompt.south) -- (1.60,5.37) -| (ar.north);
\draw[inference] (prompt.south) -- (1.60,5.37) -| (glyph.north);
\draw[inference] (ar.south) -- (.83,4.44) -| (condition.north);
\draw[inference] (glyph.south) -- (2.36,4.44) -| (condition.north);

\node[box,draw=blue,line width=1.0pt,fill=white,minimum width=4.85cm,minimum height=1.20cm] (student) at (6.55,5.35) {};
\node[font=\sffamily\bfseries\fontsize{10}{12}\selectfont,text=blue] at (6.55,5.67) {Trainable student DiT $F_\theta$};
\foreach \xx/\tt in {4.89/1000,6.00/750,7.11/500,8.22/250}{
  \node[box,draw=blue!35,fill=bluewash,minimum width=.76cm,minimum height=.31cm,inner sep=2pt,font=\sffamily\fontsize{8}{9}\selectfont] at (\xx,5.08) {$\tt$};
}
\foreach \xx in {5.41,6.52,7.63}{\draw[inference,draw=blue,line width=.55pt] (\xx,5.08) -- ++(.15,0);}
\draw[inference] (condition.east) -- (3.33,3.97) -- (3.33,5.65) -- (4.125,5.65);
\draw[inference] (noise.east) -- (3.65,3.12) -- (3.65,5.08) -- (4.125,5.08);

\node[box,draw=blue!60,fill=white,text width=1.90cm,minimum height=.55cm] (latent) at (5.18,3.97) {Latent $\hat{x}_0$};
\node[box,draw=muted!55,fill=white,text width=1.90cm,minimum height=.48cm] (vae) at (5.18,3.10) {VAE (fixed)};
\node[box,draw=blue!60,fill=white,text width=1.65cm,minimum height=.48cm] (image) at (7.80,3.10) {Image};
\node[box,draw=amber!75,fill=amberwash,text width=2.40cm,minimum height=.65cm] (pl) at (8.25,3.97) {\textbf{Local response}\\$\mathcal L_{\mathrm{PL}}$};
\draw[inference] (5.18,4.75) -- (latent.north);
\draw[inference] (latent.south) -- (vae.north);
\draw[inference] (vae.east) -- (image.west);
\draw[training,draw=amber] (8.25,4.75) -- (pl.north);

\node[box,draw=blue!55,fill=white,text width=4.75cm,minimum height=.62cm] (noised) at (13.48,5.50) {Noised student output $x_t$};
\node[box,draw=muted!55,fill=white,text width=2.50cm,minimum height=.70cm] (teacher) at (11.85,4.27) {Teacher score $s_T$\\\textcolor{muted}{Frozen}};
\node[box,draw=blue!60,fill=white,text width=2.50cm,minimum height=.70cm] (fake) at (15.08,4.27) {Fake score $s_F$\\\textcolor{blue}{Online updates}};
\node[box,draw=blue!60,fill=white,text width=5.00cm,minimum height=.55cm] (dm) at (13.48,3.18) {\textbf{Distribution matching} $\mathcal L_{\mathrm{DM}}$};
\draw[scorepath] (student.east) -- ++(.42,0) |- (noised.west);
\node[label] at (10.01,5.69) {perturb};
\draw[scorepath] (noised.south) -- ++(0,-.23) -| (teacher.north);
\draw[scorepath] (noised.south) -- ++(0,-.23) -| (fake.north);
\draw[scorepath] (teacher.south) -- ++(0,-.19) -| (dm.north);
\draw[scorepath] (fake.south) -- ++(0,-.19) -| (dm.north);

\node[box,draw=blue!45,fill=bluewash,text width=10.10cm,minimum height=.45cm] (objective) at (11.80,1.98) {\textbf{Student objective}\qquad $\mathcal L_{\mathrm{total}}=\mathcal L_{\mathrm{base}}+\lambda_{\mathrm{PL}}\mathcal L_{\mathrm{PL}}$};
\draw[training,draw=amber] (pl.east) -- (9.64,3.97) -- (9.64,2.205);
\draw[training,draw=blue] (dm.south) -- (13.48,2.205);
\node[label,anchor=west] at (.15,2.06) {Solid: forward computation};
\node[label,anchor=west] at (.15,1.80) {Dashed: training signals};

\draw[panel,draw=teal!35,fill=tealwash] (0,0) rectangle (17.0,1.48);
\node[heading,text=teal] at (.22,1.28) {D\quad Progressive teacher-guided text training};
\node[box,draw=teal!45,fill=white,text width=4.20cm,minimum height=.40cm,inner sep=3pt] (stage1) at (3.05,.78) {\textbf{Stage 1}\quad Generic prompts};
\node[box,draw=teal!45,fill=white,text width=4.20cm,minimum height=.40cm,inner sep=3pt] (stage2) at (8.50,.78) {\textbf{Stage 2}\quad Short text};
\node[box,draw=teal!45,fill=white,text width=4.20cm,minimum height=.40cm,inner sep=3pt] (stage3) at (13.95,.78) {\textbf{Stage 3}\quad Dense layouts};
\draw[inference,draw=teal] (stage1.east) -- (stage2.west);
\draw[inference,draw=teal] (stage2.east) -- (stage3.west);
\node[font=\sffamily\fontsize{8}{9}\selectfont,text=teal] at (8.50,.28) {Teacher-generated images in every GLM training stage};
\draw[training] (stage1.north) -- (3.05,1.63) -- (3.87,1.63) -- (3.87,4.60) -- (4.65,4.60) -- (4.65,4.75);
\end{tikzpicture}}
\caption{Four-step distillation with complementary response and data supervision. Frozen autoregressive and glyph encoders retain semantic and character conditions. Distribution matching transfers teacher synthesis behavior, while the path-length term regularizes the student's local response. Teacher-guided progressive training combines explicit text examples with visually integrated typography; the VAE remains fixed.}
\label{fig:pipeline}
\end{figure*}

\subsection{Training Objective}
\label{subsec:training-objective}
We augment the base distribution-matching generator objective with a path-length penalty:
\begin{equation}
\mathcal{L}_{\mathrm{total}}=\mathcal{L}_{\mathrm{base}}
+\lambda_{\mathrm{PL}}\mathcal{L}_{\mathrm{PL}}.
\label{eq:full_objective}
\end{equation}
The fake-score model is optimized separately through its score-learning objective. The additional term regularizes the student denoising output with respect to the current latent, with prior and text conditions held fixed.

\subsection{Local Response Regularization}
\label{sec:response}
Each transition of a four-step student covers a substantial portion of the denoising trajectory. Consistent local responses provide a regularity prior for this compressed mapping. Let $F_\theta(x_t,c,t)$ denote the conditional denoising output and $J=\partial F_\theta/\partial x_t$ its input Jacobian. Following path-length regularization~\citep{karras2020stylegan2}, we define
\begin{align}
r_\theta(x_t,c,t;y)&=\|J^\top y\|_2,\quad y\sim\mathcal{N}(0,I),\notag\\
\mathcal{L}_{\mathrm{PL}}&=\mathbb{E}_{x_t,c,t,y}
\left(r_\theta(x_t,c,t;y)-\operatorname{sg}(a)\right)^2,
\label{eq:lip_loss}
\end{align}
where $a$ is an exponential-moving-average response scale and $\operatorname{sg}$ denotes stop-gradient. Reverse-mode differentiation gives the vector--Jacobian product (VJP) $J^\top y=\nabla_{x_t}\langle F_\theta,y\rangle$ without materializing $J$. An input-direction Jacobian--vector product (JVP) is $Jv$. Section~\ref{sec:normalized_response} specifies the normalized statistic and target update.

For Gaussian output directions, the response energy satisfies
\begin{equation}
\mathbb{E}_y\|J^\top y\|_2^2
=\operatorname{tr}(JJ^\top)=\|J\|_F^2.
\label{eq:jacobian_moment}
\end{equation}
For a fixed target $a$, the penalty decomposes as
\begin{equation}
\mathbb{E}(r_\theta-a)^2
=\operatorname{Var}(r_\theta)+(\mathbb{E}r_\theta-a)^2.
\label{eq:response_variance}
\end{equation}
The objective regularizes response dispersion and aligns the mean with a running scale. The adaptive target encourages consistent local response magnitudes without imposing a fixed bound on the Jacobian norm.

Latent differentiation holds the AR prior and glyph embeddings fixed. Applying the penalty before classifier-free guidance targets the conditional refinement field directly.

\subsection{Normalized Responses and Adaptive Targets}
\label{sec:normalized_response}
For input and output dimensions $d_i$ and $d_o$, we normalize the output projection and use an input root-mean-square reduction:
\begin{align}
g_b&=\nabla_{x_b}\frac{\langle F_\theta(x_b,c_b,t_b),y_b\rangle}{\sqrt{d_o}},\notag\\
\widetilde r_b&=\sqrt{\frac{\|g_b\|_2^2}{d_i}+\varepsilon},
\qquad \varepsilon=10^{-8}.
\label{eq:normalized_response}
\end{align}
Each example has an independent Gaussian $y_b$. With batch size $B$, the loss and target update are
\begin{align}
\mathcal L_{\mathrm{PL},n}
&=\frac1B\sum_{b=1}^{B}
\left(\widetilde r_b-\operatorname{sg}(a_n)\right)^2,\notag\\
a_{n+1}&=\beta a_n+(1-\beta)\frac1B
\sum_{b=1}^{B}\operatorname{sg}(\widetilde r_b),
\label{eq:target_update}
\end{align}
where $\beta=0.99$. The previous target defines the current loss, and the detached batch mean updates the target for the next iteration. Per-example differentiation gives $\mathbb E_y[\widetilde r^2-\varepsilon]=\|J\|_F^2/(d_i d_o)$ for mappings without cross-sample coupling. The GLM coefficient sweep uses this fixed normalization.

\paragraph{Response consistency versus derivative matching.}
A Jacobian-energy penalty suppresses aggregate response magnitude through $\mathbb E_y\|J^\top y\|_2^2$. Our loss instead maintains consistency around an adaptive scale. GAD aligns the directional responses of teacher and student mappings~\citep{huang2026gad}; our added term uses the scalar magnitude of the student VJP. Optimizing this penalty requires higher-order differentiation during training; the deployed denoiser retains the original inference structure.

\paragraph{Response propagation in coarse transitions.}
For $\Psi(x)=x+\Delta t F_\theta(x,c,t)$ with fixed conditions,
\begin{equation}
\Psi(x+\delta x)-\Psi(x)
=(I+\Delta t J)\delta x+o(\|\delta x\|_2).
\label{eq:transition_response}
\end{equation}
The interval $\Delta t$ scales the Jacobian contribution to local propagation. With four transitions, each update carries a substantial part of scene refinement. Response regularization supplies a local prior for these coarse updates, while distribution matching and text supervision determine the target appearance and content. Section~\ref{sec:pl_ablation} evaluates the resulting quality balance across regularization weights.

\subsection{Progressive Text Training}
\label{sec:text_training}
Text training follows the three-stage curriculum in Table~\ref{tab:curriculum}: generic scenes, short strings, and dense layouts. Teacher-generated images remain present throughout to retain scene appearance and typography. Synthetic examples introduce exact strings and controlled layouts in the second stage.

The second-stage batch distribution is
\begin{equation}
P_2=(1-\pi)P_{\mathrm{image}}+\pi P_{\mathrm{text\ batch}},
\label{eq:data_mix}
\end{equation}
where $P_{\mathrm{image}}$ denotes the teacher-image stream and $P_{\mathrm{text\ batch}}$ denotes the stream containing synthetic text examples. We insert one text-stream batch after every five image-stream batches, giving $\pi=1/6$. Text-stream batches also contain teacher-generated images; Section~\ref{sec:text_construction} specifies the sample composition.

\begin{table}[t]
\centering
\scriptsize
\setlength{\tabcolsep}{3pt}
\caption{Three-stage text-training curriculum for \glm{}. Teacher images preserve scene appearance throughout, while synthetic text supervision progresses from short strings to dense layouts.}
\label{tab:curriculum}
\begin{tabular}{p{.075\columnwidth}p{.455\columnwidth}p{.35\columnwidth}}
\toprule
Stage & Training examples & Primary purpose\\
\midrule
1 & Generic prompts and teacher images & Preserve scene fidelity\\
2 & Synthetic short text and teacher images & Learn glyph-level alignment\\
3 & Dense layouts and teacher images & Refine multi-region text\\
\bottomrule
\end{tabular}
\end{table}

Synthetic examples provide exact character targets; teacher images provide scene-integrated typography. The \qwen{} study uses a teacher-specific synthetic-data construction procedure~\citep{wu2025qwenimage} to evaluate text-oriented distillation with a second conditioning architecture.

\subsection{Text-Data Construction and Stream Composition}
\label{sec:text_construction}
Generic prompts come from DiffusionDB~\citep{wang2023diffusiondb}, with exact duplicates removed after whitespace deletion. Synthetic examples place Chinese or English strings on whiteboards, billboards, blackboards, and screens. Vocabulary items supply short text; longer strings and multiple regions provide sentence- and layout-level supervision. Teacher-generated images complement these examples with scene-integrated perspective, material appearance, illumination, and typography.

The curriculum first establishes generic scene synthesis, then introduces short text, and finally emphasizes longer strings and dense layouts. Teacher imagery remains present throughout, connecting explicit character supervision to the teacher's broader visual distribution.

In stage 2, one text-stream batch follows five image-stream batches. Each image batch contains 128 teacher images; each text batch combines 64 synthetic examples and 64 teacher images. A six-batch cycle therefore contains 704 teacher images and 64 synthetic examples, giving
\begin{equation}
P_{2,\mathrm{sample}}=\frac{11}{12}P_T+\frac{1}{12}P_S,
\label{eq:sample_mix}
\end{equation}
where $P_T$ and $P_S$ are teacher-image and synthetic-text distributions. The $5{:}1$ batch-stream schedule thus corresponds to an $11{:}1$ sample mixture under this composition. This mixture combines explicit character targets with examples of typography integrated into natural scenes.

\section{Experiments}
\label{sec:experiments}
\subsection{Experimental Setup}
\paragraph{Models and training.}
We use \glm{} as the primary teacher and \qwen{} for a second-model evaluation. Teachers use 50 denoising steps, while students use four at $1024\times1024$ resolution. Generic distillation draws prompts from DiffusionDB~\citep{wang2023diffusiondb}. Training uses AdamW, four fake-score updates per generator update, and a path-length coefficient of $10^{-5}$. Text-oriented training follows the three-stage schedule in Section~\ref{sec:text_training}. Sections~\ref{sec:protocol} and~\ref{sec:text_construction} specify optimization and data construction.

\paragraph{Benchmarks and metrics.}
GenEval~\citep{ghosh2023geneval} evaluates compositional alignment across Single Object, Two Objects, Counting, Colors, Position, and Attribute Binding, together with Overall. CVTG-2K~\citep{du2025textcrafter} evaluates rendered text with word accuracy (WA), normalized edit similarity (NED), and CLIPScore. WA-$k$ denotes the group of images with $k$ text regions, for $k\in\{2,3,4,5\}$. 

For a target string $s$ and OCR output $\hat{s}$, we use the higher-is-better NED convention
\begin{equation}
\operatorname{NED}(s,\hat{s})
=1-\frac{\operatorname{Lev}(s,\hat{s})}
{\max(|s|,|\hat{s}|)},
\label{eq:ned}
\end{equation}
where $\operatorname{Lev}$ is Levenshtein distance; two empty strings have similarity 1. CLIPScore measures image--text representation alignment. The auxiliary VAE study reports reconstruction FID and PSNR, latent scale factor, and empirical response ratios. Downstream EDM2 generation is evaluated with FID and FD-DINOv2.

\paragraph{Component comparisons.}
We compare the full student with TDM~\citep{luo2025tdm} on \glm{} and with variants that omit path-length regularization or text-oriented training. The \glm{} study also compares synthetic-only and teacher-image-mixed text training and varies the PL coefficient over $\{10^{-6},10^{-5},10^{-4}\}$. The distilled models use a four-step inference budget. Table~\ref{tab:controls} summarizes the settings.
\begin{table}[t]
\centering
\scriptsize
\setlength{\tabcolsep}{3pt}
\caption{Shared distillation and evaluation settings. Each student uses four denoising transitions while retaining the teacher's conditioning and decoding interfaces.}
\label{tab:controls}
\begin{tabular}{ll}
\toprule
Component & Setting\\
\midrule
Teachers & \glm{} and \qwen{}\\
Sampling budget & Teacher: 50 steps; student: four steps\\
Resolution & $1024\times1024$\\
Student time labels & 1000, 750, 500, 250\\
Update ratio & Four fake-score updates per generator update\\
Response penalty & Enabled or disabled\\
Text training & None, synthetic-only, or teacher-image-mixed\\
GLM batch streams & Five image batches per text batch\\
\bottomrule
\end{tabular}
\end{table}

At inference, prior tokens and glyph embeddings are computed once and held fixed throughout the four scheduler transitions. With classifier-free guidance, each transition combines conditional and unconditional predictions. Step counts refer to scheduler transitions; retained-module costs are additional to the denoising loop.

\begin{figure}[t]
\centering
\scriptsize
\setlength{\tabcolsep}{2pt}
\begin{tabular}{c c}
\includegraphics[width=0.43\columnwidth]{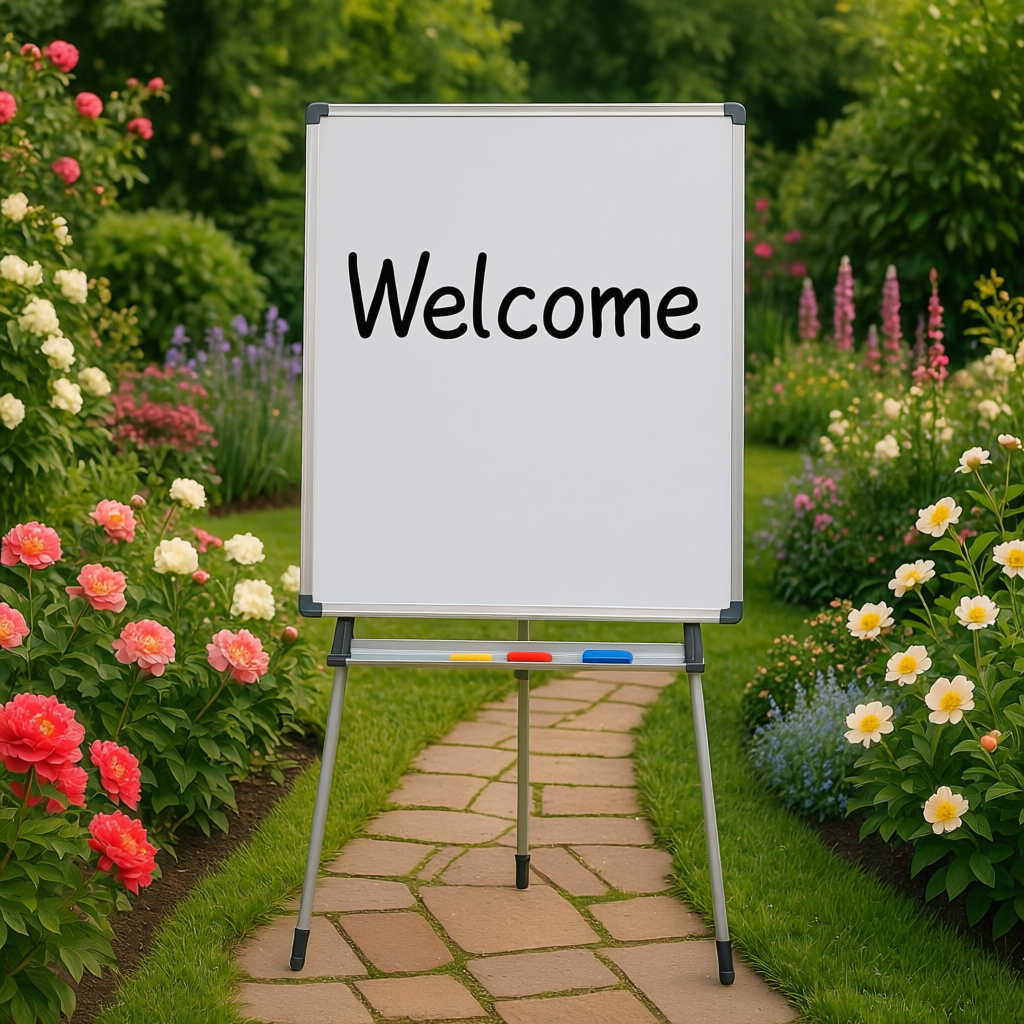} &
\includegraphics[width=0.43\columnwidth]{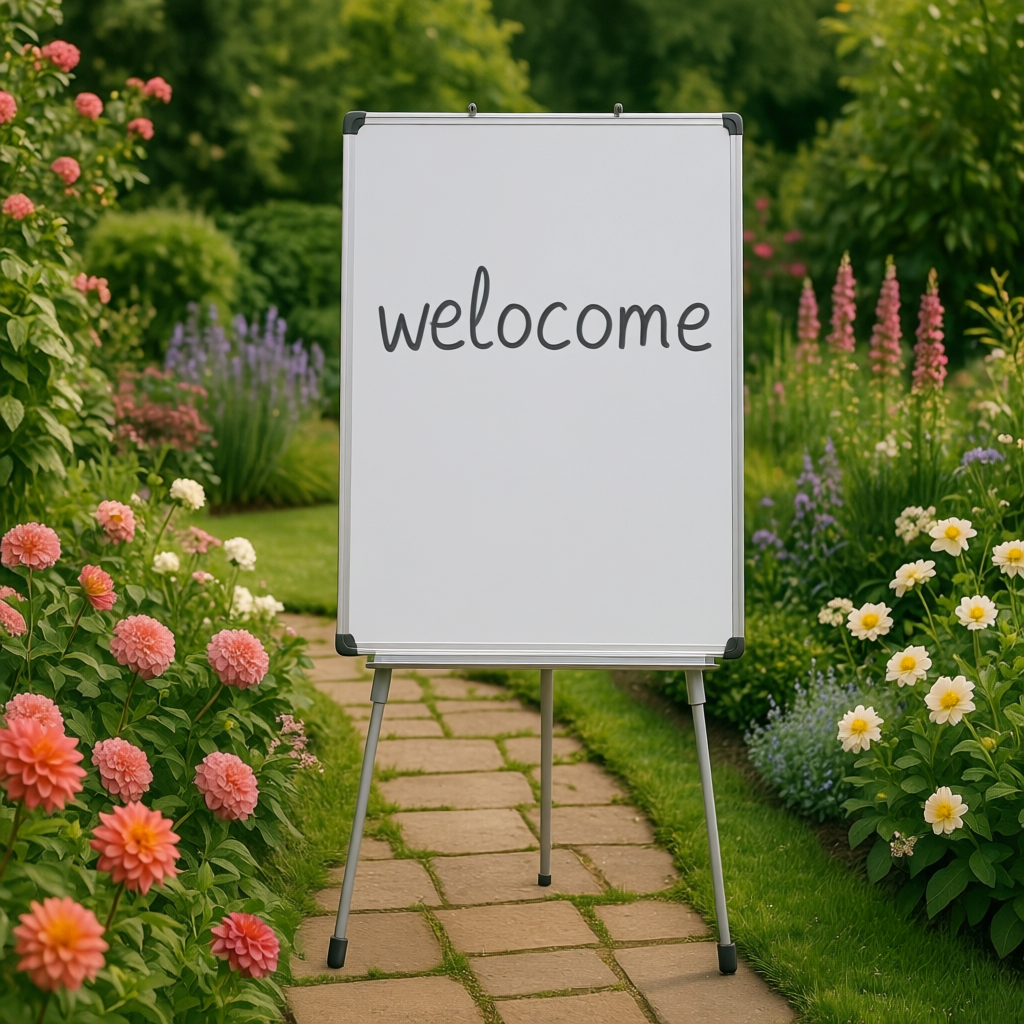}\\[-1pt]
\parbox[c]{0.43\columnwidth}{\centering\scriptsize Teacher (50 steps)} &
\parbox[c]{0.43\columnwidth}{\centering\scriptsize Ours w/o text training (4 steps)}\\[4pt]
\includegraphics[width=0.43\columnwidth]{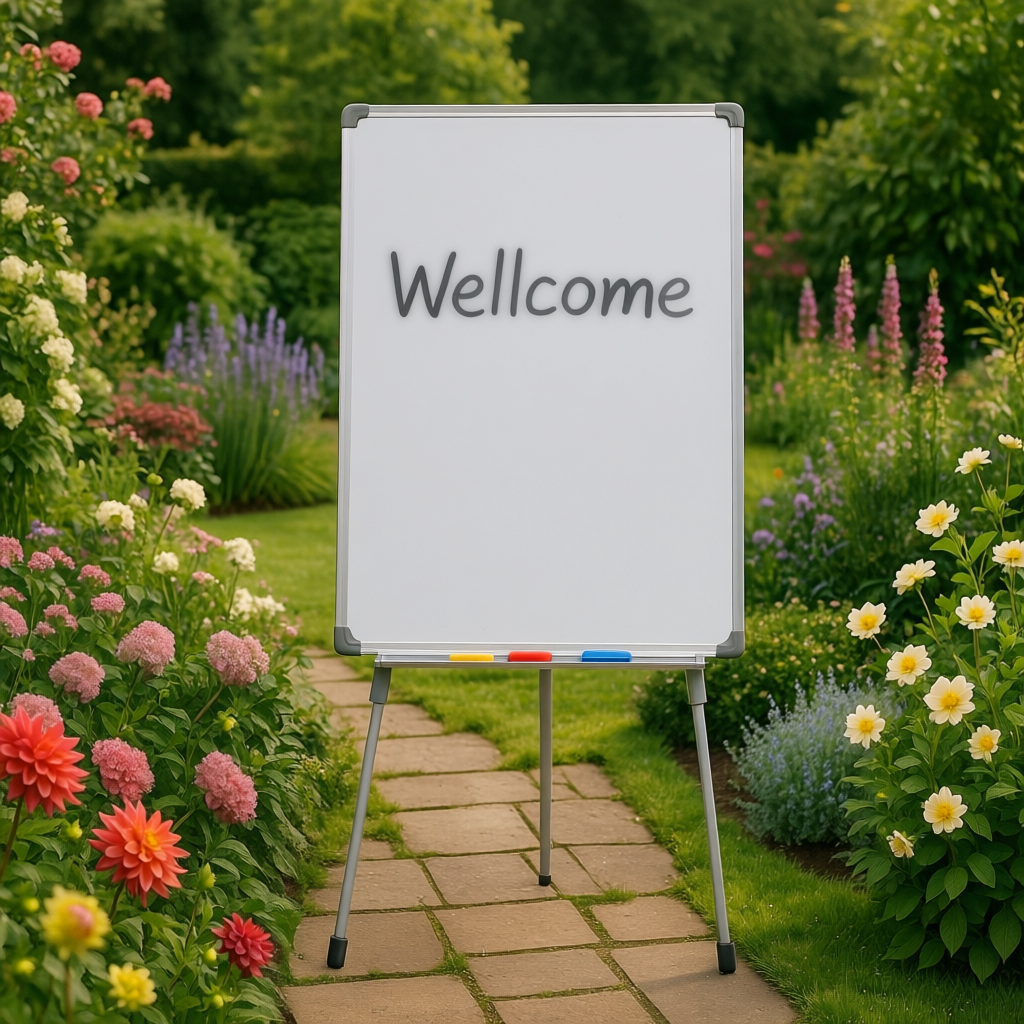} &
\includegraphics[width=0.43\columnwidth]{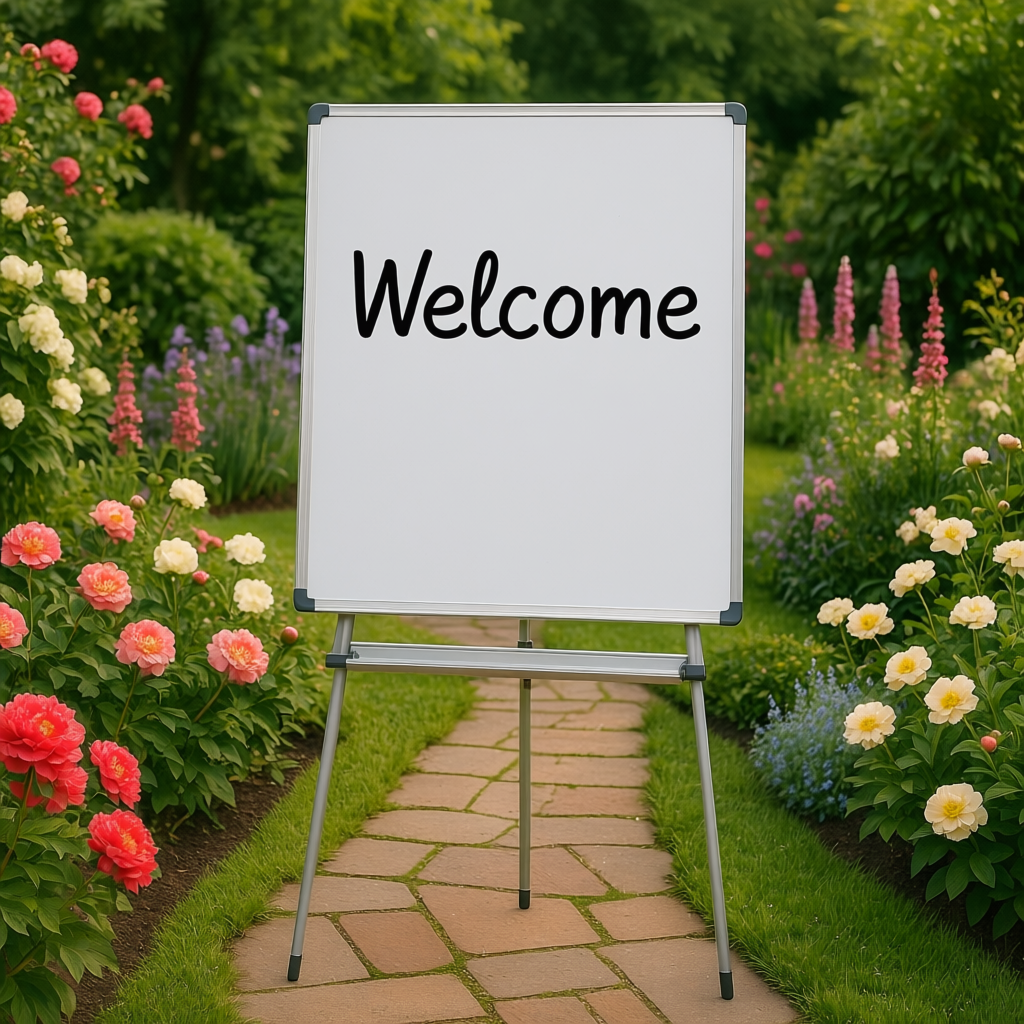}\\[-1pt]
\parbox[c]{0.43\columnwidth}{\centering\scriptsize Ours with synthetic text only (4 steps)} &
\parbox[c]{0.43\columnwidth}{\centering\scriptsize Ours with synthetic text and teacher images (4 steps)}\\
\end{tabular}
\caption{Literal text preservation under a shared garden prompt and random seed. The full student renders ``Welcome'' correctly in four denoising steps, matching the target spelling of the 50-step teacher. Intermediate variants show generation without text training and with synthetic text alone.}
\label{fig:qualitative}
\end{figure}

\subsection{Optimization and Inference Protocol}
\label{sec:protocol}
Table~\ref{tab:optimization} lists optimization settings. The student combines distribution-matching and response losses; the online fake-score branch receives four updates per generator update. AR, glyph, and VAE modules remain fixed.
\begin{table}[t]
\centering\small
\setlength{\tabcolsep}{4pt}
\caption{Optimization settings. Response regularization adds a training term while preserving the student's inference architecture.}
\label{tab:optimization}
\begin{tabular}{ll}
\toprule
Setting & Value\\
\midrule
Optimizer & AdamW\\
Learning rate & $10^{-6}$\\
Betas; weight decay & $(0.9,0.999)$; 0.01\\
Batch size & 16/GPU, 8 GPUs, total 128\\
PL target decay & 0.99\\
Default PL coefficient & $10^{-5}$\\
CFG scale & 4.0\\
Evaluation seed & 42\\
Checkpoint selection & Final checkpoint\\
\bottomrule
\end{tabular}
\end{table}

Student timestep labels are 1000, 750, 500, and 250; the teacher-specific scheduler supplies noise levels and model-input indices. Prior tokens and glyph embeddings are computed once and reused, and the terminal latent is decoded once.

\subsection{Inference Efficiency}
\label{sec:efficiency}
With guidance scale $\gamma$, each transition combines
\begin{equation}
F_{\mathrm{CFG}}=F_{\mathrm{uncond}}+\gamma
(F_{\mathrm{cond}}-F_{\mathrm{uncond}}).
\label{eq:cfg}
\end{equation}
Table~\ref{tab:inference_budget} reports scheduler transitions and conditional/unconditional branch evaluations. Guidance branches can share a batched invocation. The response penalty is evaluated only during training.
\begin{table}[t]
\centering\footnotesize
\setlength{\tabcolsep}{3pt}
\caption{Scheduled denoising workload under classifier-free guidance. Reductions apply to branch-evaluation counts; total latency also includes conditioning and VAE decoding.}
\label{tab:inference_budget}
\begin{tabular}{lrrr}
\toprule
Model & Steps & Branch evals. & Reduction\\
\midrule
Teacher & 50 & 100 & $1\times$\\
TDM & 4 & 8 & $12.5\times$\\
Ours & 4 & 8 & $12.5\times$\\
\bottomrule
\end{tabular}
\end{table}
Writing retained-module time as $T_{\mathrm{fixed}}$ and guided-transition time as $T_{\mathrm{step}}$, a simple cost model gives
\begin{equation}
\frac{T(50)}{T(4)}
=\frac{T_{\mathrm{fixed}}+50T_{\mathrm{step}}}
{T_{\mathrm{fixed}}+4T_{\mathrm{step}}}.
\label{eq:cost_model}
\end{equation}
Table~\ref{tab:measured_efficiency} reports measured single-image inference cost for the GLM teacher and our student on an NVIDIA A100 GPU. Timing covers the generation pipeline, including autoregressive generation and text encoding, and is recorded with Python\texttt{ tqdm}. GPU memory usage is recorded with an external monitoring tool. Latency decreases from 5.818\,s to 1.742\,s, corresponding to a $3.34\times$ speedup and a 70.1\% reduction in generation time. Both models require 40\,GB of GPU memory. In this implementation, DMD compresses the denoising trajectory while retaining the model architecture, conditioning modules, and decoder; it does not compress parameter storage. The measured benefit is therefore faster inference at the same memory requirement.

\begin{table}[t]
\centering\footnotesize
\setlength{\tabcolsep}{4pt}
\caption{Measured single-image inference cost for \glm{} on an NVIDIA A100 GPU. Our four-step student reduces latency by 70.1\% while retaining the teacher's 40\,GB GPU memory requirement.}
\label{tab:measured_efficiency}
\begin{tabular}{lrrr}
\toprule
Model & Time (s/image) & Memory (GB) & Speedup\\
\midrule
Teacher (50 steps) & 5.818 & 40 & $1.00\times$\\
Ours (4 steps) & 1.742 & 40 & $3.34\times$\\
\bottomrule
\end{tabular}
\end{table}

The $12.5\times$ reduction in denoising evaluations is a workload ratio, distinct from the measured $3.34\times$ latency speedup. Retained conditioning and decoding costs limit the acceleration predicted by Eq.~\ref{eq:cost_model}. At the shared four-step budget, our student also improves composition and text fidelity over TDM.

\subsection{Interpreting Visual and Textual Evaluation}
\label{sec:metric_interpretation}
GenEval tests objects, counts, colors, and spatial or attribute relations; CVTG-2K tests literal strings within visual compositions. Exact WA is sensitive to substitutions, insertions, and omissions, while NED grades partial character agreement and CLIPScore measures broader image--text compatibility. WA-$k$ groups images by their number of text regions. The qualitative comparison uses a common prompt and seed, pairing complete scenes with the identically positioned text crops in Figure~\ref{fig:teaser}.

\subsection{Main Results on GLM-Image}
\begin{table*}[t]
\centering
\footnotesize
\setlength{\tabcolsep}{2.5pt}
\caption{Text-rendering fidelity on CVTG-2K with \glm{}. Teacher-image mixing achieves the strongest four-step result on every metric, outperforming TDM and both text-training variants. The teacher uses 50 steps. Higher is better; boldface denotes the best four-step score.}
\label{tab:glm_text}
\begin{tabular*}{\textwidth}{@{}l@{\extracolsep{\fill}}ccccccc@{}}
\toprule
Method & WA-2$\uparrow$ & WA-3$\uparrow$ & WA-4$\uparrow$ & WA-5$\uparrow$ & WA$\uparrow$ & NED$\uparrow$ & CLIPScore$\uparrow$\\
\midrule
Teacher & 0.9103 & 0.9209 & 0.9169 & 0.8975 & 0.9116 & 0.9557 & 0.7877\\
TDM~\citep{luo2025tdm} & 0.2179 & 0.2088 & 0.1300 & 0.1194 & 0.1690 & 0.4265 & 0.6902\\
Student w/o text training & 0.2032 & 0.1758 & 0.1087 & 0.0951 & 0.1457 & 0.3717 & 0.6895\\
Student with synthetic text & 0.2448 & 0.2298 & 0.1910 & 0.1691 & 0.2086 & 0.5050 & 0.7283\\
Student with teacher-image mixing & \textbf{0.5548} & \textbf{0.5744} & \textbf{0.4868} & \textbf{0.4022} & \textbf{0.5045} & \textbf{0.7130} & \textbf{0.7508}\\
\bottomrule
\end{tabular*}
\end{table*}

\begin{table*}[t]
\centering
\footnotesize
\setlength{\tabcolsep}{2.5pt}
\caption{Text-rendering fidelity on CVTG-2K with \qwen{}. Dedicated text training improves all region-count groups, WA, NED, and CLIPScore. The teacher uses 50 steps and students use four. Higher is better; boldface denotes the best student score.}
\label{tab:qwen_text}
\begin{tabular*}{\textwidth}{@{}l@{\extracolsep{\fill}}ccccccc@{}}
\toprule
Method & WA-2$\uparrow$ & WA-3$\uparrow$ & WA-4$\uparrow$ & WA-5$\uparrow$ & WA$\uparrow$ & NED$\uparrow$ & CLIPScore$\uparrow$\\
\midrule
Teacher & 0.8370 & 0.8364 & 0.8313 & 0.8158 & 0.8288 & 0.9116 & 0.8017\\
Student w/o text training & 0.4684 & 0.3576 & 0.3203 & 0.2751 & 0.3554 & 0.4520 & 0.7785\\
Full student & \textbf{0.6878} & \textbf{0.6625} & \textbf{0.6217} & \textbf{0.5860} & \textbf{0.6395} & \textbf{0.8542} & \textbf{0.7873}\\
\bottomrule
\end{tabular*}
\end{table*}

The full four-step student achieves WA 0.5045, NED 0.7130, and CLIPScore 0.7508 (Table~\ref{tab:glm_text}), improving over TDM by 0.3355, 0.2865, and 0.0606. Gains occur in every region-count group. For five-region scenes, WA rises from 0.1194 to 0.4022, showing improved preservation of multiple strings within one composition.

Teacher-image mixing raises WA from 0.2086 to 0.5045 over synthetic-only training and improves NED and CLIPScore. Synthetic-only training also improves on the student without text supervision. These comparisons support combining exact string targets with examples of realistic scene typography. Figure~\ref{fig:qualitative} shows the corresponding rendering behavior under a shared prompt and seed.

\begin{table}[t]
\centering
\scriptsize
\setlength{\tabcolsep}{2.5pt}
\caption{Compositional fidelity on \glm{}. Our four-step student improves all six GenEval categories over four-step TDM, raising Overall from 0.81 to 0.84. The teacher uses 50 steps. Higher is better; boldface marks the best distilled-model score.}
\label{tab:glm_geneval}
\resizebox{\columnwidth}{!}{%
\begin{tabular}{lccccccc}
\toprule
Method & Single & Two & Count & Color & Position & Attr. & Overall\\
\midrule
Teacher & 0.99 & 0.93 & 0.89 & 0.90 & 0.78 & 0.77 & 0.88\\
TDM~\citep{luo2025tdm} & 0.97 & 0.88 & 0.80 & 0.80 & 0.72 & 0.69 & 0.81\\
Student w/o PL & 0.97 & 0.88 & 0.81 & 0.82 & 0.71 & 0.68 & 0.81\\
Full student & \textbf{0.98} & \textbf{0.90} & \textbf{0.84} & \textbf{0.85} & \textbf{0.73} & \textbf{0.72} & \textbf{0.84}\\
\bottomrule
\end{tabular}
}
\end{table}

\paragraph{Compositional fidelity.}
The full student reaches 0.84 GenEval Overall, compared with 0.81 for TDM and the student without PL (Table~\ref{tab:glm_geneval}). Improvements over TDM cover all six categories, including 0.04 in Counting, 0.05 in Colors, and 0.03 in Attribute Binding. The PL ablation further supports its role in preserving object relations and attribute constraints.

\paragraph{Multi-region text.}
Teacher-image mixing improves WA over synthetic-only training by 0.3100, 0.3446, 0.2958, and 0.2331 for two through five regions. NED improves concurrently, indicating better character-level agreement. Together, these results show that teacher imagery strengthens literal rendering across both sparse and dense text layouts.

\subsection{Ablation on Regularization Strength}
\label{sec:pl_ablation}
The intermediate weight $\lambda_{\mathrm{PL}}=10^{-5}$ gives the best score on all four metrics (Table~\ref{tab:pl_weight}): WA 0.5045, NED 0.7130, CLIPScore 0.7508, and GenEval Overall 0.84. The larger weight $10^{-4}$ yields WA 0.4648 and Overall 0.83, while $10^{-6}$ yields 0.4475 and 0.82. NED and CLIPScore follow the same ranking. Among the tested coefficients, moderate regularization best preserves both character detail and scene composition.
\begin{table}[t]
\centering
\small
\setlength{\tabcolsep}{4pt}
\caption{PL-weight ablation on \glm{} after text-oriented training. The intermediate coefficient $10^{-5}$ gives the best result on all four metrics. Every setting uses four-step inference. Higher is better; boldface marks the best score.}
\label{tab:pl_weight}
\begin{tabular}{lcccc}
\toprule
$\lambda_{\mathrm{PL}}$ & WA & NED & CLIPScore & GenEval\\
\midrule
$10^{-6}$ & 0.4475 & 0.6821 & 0.7404 & 0.82\\
$10^{-5}$ & \textbf{0.5045} & \textbf{0.7130} & \textbf{0.7508} & \textbf{0.84}\\
$10^{-4}$ & 0.4648 & 0.6997 & 0.7460 & 0.83\\
\bottomrule
\end{tabular}
\end{table}

\subsection{Cross-Model Evaluation on Qwen-Image}

\begin{table}[t]
\centering
\scriptsize
\setlength{\tabcolsep}{2.5pt}
\caption{Effect of response regularization on \qwen{}. PL improves four GenEval categories and raises Overall from 0.82 to 0.83. The teacher uses 50 steps and students use four. Higher is better; boldface marks the best student score, including ties.}
\label{tab:qwen_geneval}
\resizebox{\columnwidth}{!}{%
\begin{tabular}{lccccccc}
\toprule
Method & Single & Two & Count & Color & Position & Attr. & Overall\\
\midrule
Teacher & 0.99 & 0.92 & 0.89 & 0.88 & 0.76 & 0.77 & 0.87\\
Student w/o PL & \textbf{0.98} & 0.88 & 0.83 & \textbf{0.84} & 0.72 & 0.70 & 0.82\\
Full student & \textbf{0.98} & \textbf{0.89} & \textbf{0.84} & \textbf{0.84} & \textbf{0.73} & \textbf{0.71} & \textbf{0.83}\\
\bottomrule
\end{tabular}
}
\end{table}

On \qwen{}, response regularization raises GenEval Overall from 0.82 to 0.83, with gains of 0.01 in Two Objects, Counting, Position, and Attribute Binding (Table~\ref{tab:qwen_geneval}). Single Object and Colors retain their baseline scores.

Dedicated text training raises \qwen{} WA from 0.3554 to 0.6395 and NED from 0.4520 to 0.8542 (Table~\ref{tab:qwen_text}). Region-wise WA gains range from 0.2194 to 0.3109. Both teacher studies therefore show improved multi-string rendering with dedicated text supervision, using teacher-image mixing for GLM and synthetic text examples for Qwen.

\subsection{Text Complexity and Capability Preservation}
\label{sec:complexity}
\begin{figure*}[t]
\centering
\includegraphics[width=.96\textwidth]{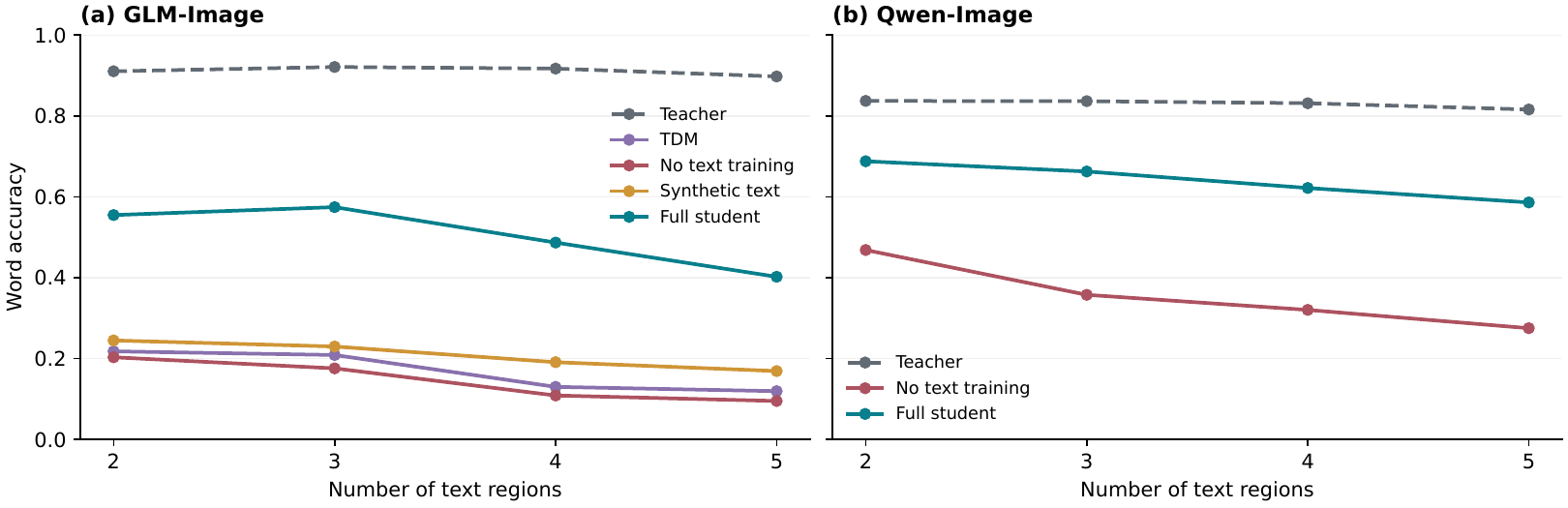}
\caption{CVTG-2K word accuracy across text-region counts. Teacher-image mixing gives the strongest GLM student results and exceeds four-step TDM in every group; dedicated text training also improves Qwen rendering throughout. Curves connect the measurements in Tables~\ref{tab:glm_text} and~\ref{tab:qwen_text}.}
\label{fig:region_curves}
\end{figure*}

Figure~\ref{fig:region_curves} summarizes accuracy across region counts. Teacher-image mixing gives the strongest GLM student performance in every group, including five-region WA 0.4022 versus 0.1691 for synthetic-only training and 0.1194 for TDM. Dedicated Qwen text training raises five-region WA from 0.2751 to 0.5860, extending the benefit to a second teacher family.

\paragraph{Scene fidelity and literal correctness.}
The full GLM student improves both composition and literal rendering at the same four-step budget as TDM. In Figure~\ref{fig:qualitative}, teacher-image mixing recovers \emph{Welcome} while preserving its garden setting. The scene and text are learned as a joint visual output.

\paragraph{Response strength and text detail.}
The intermediate coefficient yields the strongest WA, NED, CLIPScore, and GenEval results. The consistent ranking across these metrics supports balancing local response regularity with sensitivity to fine text and scene conditions.

\subsection{Supporting VAE and EDM2 Study}
\begin{table}[t]
\centering
\scriptsize
\caption{Auxiliary VAE reconstruction and downstream EDM2 generation. Response regularization improves reconstruction FID and both generation distances. Lower FID and FD-DINOv2 and higher PSNR indicate better quality. Scale denotes the latent scale factor; response ratios describe empirical sensitivity variation.}
\label{tab:supporting}
\begin{tabular}{lcccc}
\toprule
VAE & FID & PSNR & Scale & Resp. ratio\\
\midrule
Without regularization & 1.68 & 25.480 & 0.248 & 42.159\\
With regularization & 1.65 & 25.495 & 0.323 & 30.629\\
SD3 VAE reference & 1.69 & 25.784 & 0.209 & 38.167\\
\midrule
EDM2 variant & FID & FD-DINOv2 & &\\
Baseline & 5.268 & 311.906 & &\\
Regularized & 4.953 & 276.889 & &\\
\bottomrule
\end{tabular}
\end{table}

The auxiliary VAE--EDM2 study tests response regularization in a latent representation model (Table~\ref{tab:supporting}). VAE reconstruction FID improves from 1.68 to 1.65, and PSNR increases from 25.480 to 25.495. EDM2 trained on the regularized representation improves FID from 5.268 to 4.953 and FD-DINOv2 from 311.906 to 276.889.

\begin{table}[t]
\centering
\scriptsize
\caption{VAE regularization-strength sweep. The $1\times$ setting minimizes reconstruction FID, while $8\times$ supports the best downstream generation in Table~\ref{tab:supporting}. Scale is the latent scale factor. Lower FID and higher PSNR indicate better reconstruction.}
\label{tab:lip_strength}
\begin{tabular}{lcccc}
\toprule
VAE & FID & PSNR & Scale & Resp. ratio\\
\midrule
Baseline & 1.68 & 25.480 & 0.248 & 42.159\\
1$\times$ PL & 1.64 & 25.078 & 0.280 & 32.833\\
8$\times$ PL & 1.65 & 25.495 & 0.323 & 30.629\\
40$\times$ PL & 1.86 & 25.088 & 0.396 & 20.020\\
\bottomrule
\end{tabular}
\end{table}

The VAE sweep identifies different operating points for reconstruction and generation (Table~\ref{tab:lip_strength}). The $1\times$ setting gives the best reconstruction FID; $8\times$ gives the best downstream EDM2 result. EDM2 $(\mathrm{FID},\mathrm{FD\text{-}DINOv2})$ is $(5.060,286.826)$, $(4.953,276.889)$, and $(6.939,333.322)$ at $1\times$, $8\times$, and $40\times$ PL, respectively. Moderate regularization produces the strongest downstream representation.

Directional response statistics in Section~\ref{sec:auxiliary_details} characterize the accompanying change in local sensitivity. Their relative spread and absolute scale describe different aspects of the representation.

\subsection{Auxiliary Protocol and Directional Responses}
\label{sec:auxiliary_details}
\begin{figure*}[t]
\centering
\input{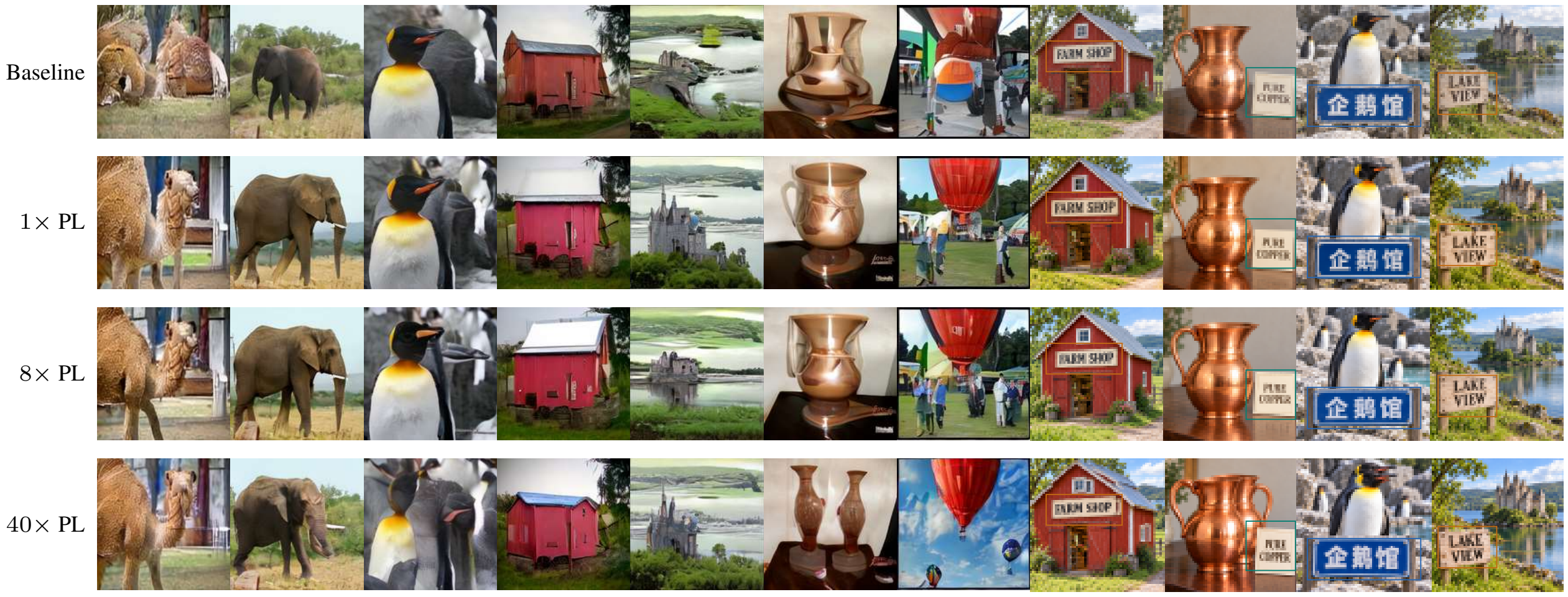}
\caption{ImageNet samples and text-scene illustrations arranged in a continuous grid. Rows correspond to the baseline and $1\times$, $8\times$, and $40\times$ PL settings. Quantitative comparisons are reported in Tables~\ref{tab:supporting} and~\ref{tab:lip_strength}.}
\label{fig:edm_samples}
\end{figure*}

\begin{figure*}[t]
\centering
\includegraphics[width=\textwidth]{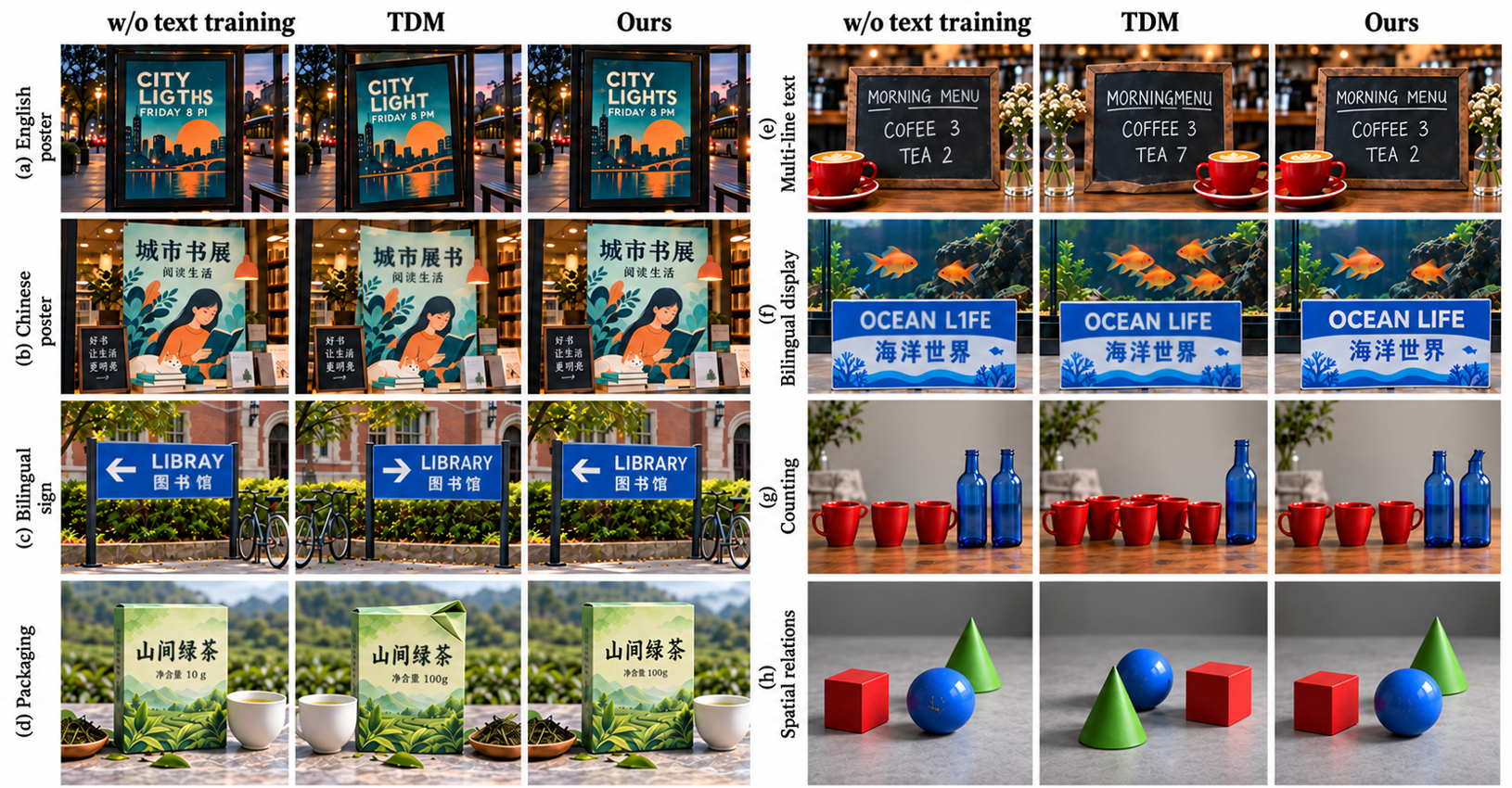}
\caption{Qualitative comparison of four-step GLM-Image students. Within each three-column group, outputs are ordered as the student without text training, TDM, and our complete framework. Cases cover (a) an English poster, (b) a Chinese poster, (c) a bilingual sign, (d) product packaging, (e) multiline text, (f) a bilingual display, (g) object counting, and (h) spatial relations. The examples illustrate differences in literal text preservation, object counts, and spatial arrangement while retaining coherent scene appearance.}
\label{fig:method_comparison}
\end{figure*}

The auxiliary study applies response regularization to a VAE trained with reconstruction, perceptual, adversarial, and KL terms. We use $\lambda_{\mathrm{adv}}=\lambda_{\mathrm{LPIPS}}=1$, $\lambda_{\mathrm{KL}}=10^{-6}$, batch size 192, and target decay 0.99. PL is evaluated once per 1536 images at $1\times$, $8\times$, and $40\times$ weights. EDM2 models trained on these representations evaluate downstream generation quality~\citep{karras2024edm2}.

To probe local behavior, consider a deterministic mapping $G$, input $u$, and normalized input direction $\delta_i$. A finite-difference response is
\begin{equation}
q_i(u;h)=\frac{\|G(u+h\delta_i)-G(u)\|_2}{h},
\qquad \|\delta_i\|_2=1.
\label{eq:finite_difference}
\end{equation}
We evaluate ten directions with $h=10^{-2}$. At small perturbations, $q_i$ approximates $\|J_G(u)\delta_i\|_2$, providing a directional sensitivity probe of the learned representation.

\begin{table}[t]
\centering\small
\caption{Directional response statistics for the auxiliary VAE sweep. Low and High report empirical finite-difference response magnitudes; Ratio is High/Low. PL narrows this relative spread, complementing the reconstruction and generation comparisons in Tables~\ref{tab:supporting} and~\ref{tab:lip_strength}.}
\label{tab:response_components}
\begin{tabular}{lrrr}
\toprule
VAE & Low & High & Ratio\\
\midrule
Baseline & 0.069 & 2.909 & 42.159\\
$1\times$ PL & 0.102 & 3.349 & 32.833\\
$8\times$ PL & 0.143 & 4.380 & 30.629\\
$40\times$ PL & 0.247 & 4.945 & 20.020\\
\bottomrule
\end{tabular}
\end{table}

Table~\ref{tab:response_components} reports the absolute response scale and its relative spread. Both Low and High increase with regularization strength, while High/Low decreases. These measurements indicate a narrower relative response profile at a larger absolute scale; the smaller ratio does not imply reduced absolute sensitivity. Reconstruction and generation metrics identify the useful operating point.

\paragraph{Reconstruction and downstream generation.}
The $1\times$ setting gives the best VAE reconstruction FID, whereas $8\times$ gives the best downstream EDM2 FID and FD-DINOv2. The strongest weight raises the latent scale factor to 0.396 and degrades downstream generation. Effective regularization balances response geometry with information preservation in the latent representation.

Figure~\ref{fig:edm_samples} pairs the original ImageNet samples with four illustrative text scenes aligned to each row's appearance. The strongest setting shows overlapping penguin contours and split vessel structure in the original samples. The additional illustrations highlight letter boundaries, line spacing, and Chinese strokes. TextInVision also examines character preservation through VAE reconstruction~\citep{fallah2025textinvision}. The measured VAE--EDM2 results favor moderate regularization.

\subsection{Qualitative Comparison across Visual Tasks}
\label{sec:qualitative_comparison}
Figure~\ref{fig:method_comparison} compares the student without text training, TDM, and our complete framework across eight cases. The text-bearing examples span English and Chinese posters, bilingual signage, product packaging, a multiline menu, and a bilingual display. They expose character omissions, substitutions, and spacing errors that can remain difficult to distinguish from overall scene plausibility. In the English poster, our student preserves ``CITY LIGHTS'' and ``FRIDAY 8 PM''; in the menu, it retains the separate lines ``COFFEE 3'' and ``TEA 2.'' The object-counting example additionally shows three red cups and two blue bottles in our output, while the spatial example places the red cube to the left of the blue sphere and the green cone behind it. These examples complement the aggregate text and compositional evaluations in Tables~\ref{tab:glm_text} and~\ref{tab:glm_geneval}.

\subsection{Discussion}
Our four-step distillation framework combines local response regularization with data tailored to literal text. Response regularization improves compositional alignment, while teacher-guided training supplies exact strings and scene-integrated typography. Their combination exceeds four-step TDM on GenEval and every CVTG-2K metric, raising WA from 0.1690 to 0.5045 at the same denoising budget.

The coefficient study connects these benefits: a moderate response penalty gives the strongest composition and text scores. Retaining the AR prior, glyph encoder, and VAE preserves the pretrained synthesis interface, so the adaptation remains concentrated on the denoiser. The response penalty adds no inference operations. Measured single-image latency falls by 70.1\%, although the retained architecture keeps the GPU memory requirement at 40\,GB.

\section{Limitations}
The present study focuses on four-step generation. Dense text layouts still leave room for improvement in character accuracy and placement. Retained conditioning and decoding modules contribute fixed deployment costs, and trajectory compression alone does not reduce the measured single-image memory requirement. Finer glyph supervision, additional sampling budgets, and joint module compression provide direct extensions of the framework.

\section{Conclusion}
We presented a four-step distillation framework for text-bearing image generation that combines local response regularization with teacher-guided text training. The method retains GLM-Image's semantic and glyph-conditioning interface while improving compositional alignment and literal rendering. It raises GenEval Overall from four-step TDM's 0.81 to 0.84 and CVTG-2K WA from 0.1690 to 0.5045. The coefficient sweep favors $10^{-5}$, and Qwen-Image experiments show further gains from response regularization and dedicated text data. The student also achieves a $3.34\times$ single-image inference speedup with the same 40\,GB GPU memory requirement as the teacher. These results support combining conditional response regularization with dedicated text supervision for efficient image synthesis.

\label{body:end}

{\small
\interlinepenalty=10000

}

\par\bigskip
\appendix
\numberwithin{equation}{section}
\section{Response Spectrum and Directional Variability}
Let the denoiser Jacobian have singular value decomposition $J=U\Sigma V^\top$. For $y\sim\mathcal N(0,I)$, the coordinates of $U^\top y$ are independent standard Gaussians. Writing these coordinates as $\xi_i$ gives
\begin{equation}
\|J^\top y\|_2^2\overset{d}{=}\sum_i\sigma_i^2\xi_i^2.
\label{eq:supp_spectral_response}
\end{equation}
Using $\operatorname{Var}(\xi_i^2)=2$ and independence,
\begin{equation}
\operatorname{Var}_y(\|J^\top y\|_2^2)=2\sum_i\sigma_i^4.
\label{eq:supp_response_variance}
\end{equation}
At fixed total squared singular value, concentrating energy in fewer directions increases squared-response variability. This relation characterizes the spectrum sampled by random projections. The main-paper loss applies to response magnitudes and averages over latent states, prompts, and timesteps.

\section{Teacher-Referenced Derivative Alignment}
For teacher and student mappings with a common input and output space, let $A=J_\theta-J_T$ and $v\sim\mathcal N(0,I)$. The directional alignment objective used in geometry-aware distillation~\citep{huang2026gad} satisfies
\begin{align}
\mathbb E_v\|Av\|_2^2
&=\operatorname{tr}(A^\top A\,\mathbb E[vv^\top])\notag\\
&=\|J_\theta-J_T\|_F^2.
\label{eq:supp_derivative_alignment}
\end{align}
The teacher supplies a vector-valued derivative target. Our adaptive response prior instead uses the scalar magnitude of a student VJP. This identity makes the target-level distinction precise while leaving the distribution-matching backbone unchanged.

\section{Coefficient Scaling under Dimension Normalization}
Let $c=d_i d_o$ be the product of input and output dimensions. For the unstabilized response, set
\begin{equation}
r_{\mathrm{norm}}=\frac{r}{\sqrt c},\qquad
a_{\mathrm{norm}}=\frac{a}{\sqrt c}.
\label{eq:supp_scaled_response}
\end{equation}
With consistently scaled targets, the corresponding losses satisfy
\begin{equation}
\mathbb E(r_{\mathrm{norm}}-a_{\mathrm{norm}})^2
=\frac1c\,\mathbb E(r-a)^2.
\label{eq:supp_loss_scaling}
\end{equation}
Thus, an equivalent coefficient in the normalized objective is $\lambda_{\mathrm{norm}}=c\lambda_{\mathrm{raw}}$. The coefficient convention is tied to the response normalization and latent dimensions. Holding these choices fixed within the GLM sweep gives a consistent comparison of regularization strength.

\end{document}